\pdfoutput=1

\documentclass[11pt]{article}

\usepackage[final]{acl}

\usepackage[utf8]{inputenc}
\usepackage[T1]{fontenc}
\usepackage{microtype}
\usepackage{inconsolata}
\usepackage{times}
\usepackage{latexsym}
\usepackage{amsmath}
\usepackage{algorithm}
\usepackage{algpseudocode}
\usepackage{natbib}
\usepackage{graphicx}
\usepackage{booktabs}
\usepackage{multirow}
\usepackage{adjustbox}
\usepackage{tabularx}
\usepackage{array}
\usepackage{color}
\usepackage{xcolor}
\usepackage{soul} 
\usepackage[normalem]{ulem}
\usepackage{colortbl}
\usepackage{longtable}
\useunder{\uline}{\ul}{}

\title{Incorporating Domain Knowledge into Materials Tokenization}

\author{
  \textbf{Yerim Oh\textsuperscript{1}} \quad
  \textbf{Jun-Hyung Park\textsuperscript{2}} \quad
  \textbf{Junho Kim\textsuperscript{1}} \quad
  \textbf{SungHo Kim\textsuperscript{1}} \quad
  \textbf{SangKeun Lee\textsuperscript{1,3}}
\\
  \textsuperscript{1}Department of Artificial Intelligence, Korea University
\\
  \textsuperscript{2}Division of Language \& AI, Hankuk University of Foreign Studies
\\
  \textsuperscript{3}Department of Computer Science and Engineering, Korea University
\\
  \texttt{\{yerim0210, monocrat, sungho3268, yalphy\}@korea.ac.kr, jhp@hufs.ac.kr}}

\begin{document}
\maketitle
\begin{abstract}

While language models are increasingly utilized in materials science, typical models rely on frequency-centric tokenization methods originally developed for natural language processing. However, these methods frequently produce excessive fragmentation and semantic loss, failing to maintain the structural and semantic integrity of material concepts. To address this issue, we propose MATTER, a novel tokenization approach that integrates material knowledge into tokenization. Based on MatDetector trained on our materials knowledge base and a re-ranking method prioritizing material concepts in token merging, MATTER maintains the structural integrity of identified material concepts and prevents fragmentation during tokenization, ensuring their semantic meaning remains intact. The experimental results demonstrate that MATTER outperforms existing tokenization methods, achieving an average performance gain of 4\% and 2\% in the generation and classification tasks, respectively. These results underscore the importance of domain knowledge for tokenization strategies in scientific text processing.\footnote{Our code is available at \url{https://github.com/yerimoh/MATTER}}
\end{abstract}

\section{Introduction}

Recent advances in language models have expanded their applications in materials science \cite{ pilania2021machine, olivetti2020data}. However, typical language models for materials science utilize frequency-centric subword tokenization methods originally developed for general natural language processing (NLP) tasks \cite{trewartha2022quantifying, gupta2022matscibert, huang2022batterybert}. These methods prioritize high-frequency words in tokenization, resulting in misrepresentation of low-frequency words \cite{yuan2024vocabulary, lee2024length, liang2023xlm}, which is particularly problematic in material corpora. 

\begin{figure}[t] 
    \centering 
    \includegraphics[width=0.48\textwidth]{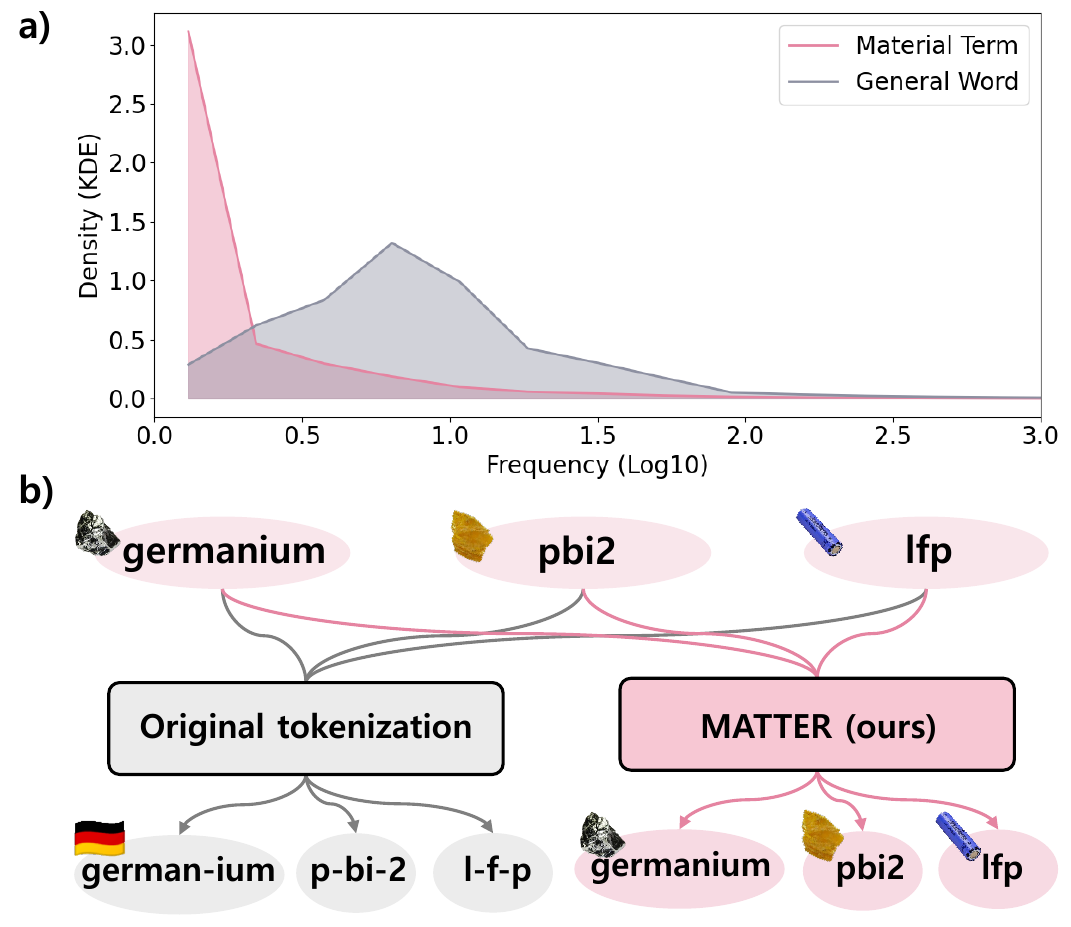} 
    \caption{(a) Frequency histograms of material concepts and general words on 150K materials-related scientific papers. (b) Tokenization results of material concepts using conventional tokenization and MATTER (ours).} 
    \label{fig:1_graph} 
\end{figure}

Material concepts—such as material names and chemical formulas—tend to appear infrequently in materials-related scientific papers as shown in Figure \ref{fig:1_graph}(a). This can lead to the oversight of material concepts in frequency-centric tokenization methods, whereas high-frequency general words dominate the subword vocabulary. As a result, material concepts are indeed fragmented into semantically unrelated subwords. For example, as shown in Figure 1(b), the word \emph{germanium}\includegraphics[width=3.5mm]{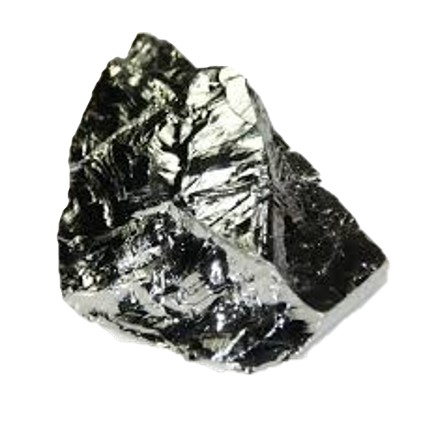}, which means a chemical element, is split into semantically unrelated subwords \emph{german}\includegraphics[width=4.5mm]{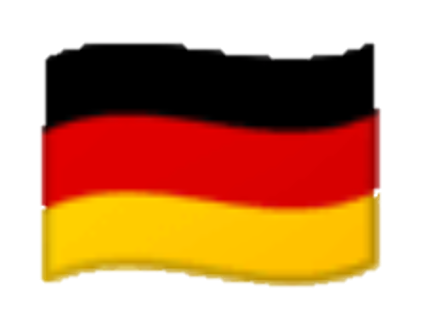} and \emph{-ium}. Such fragmentation may cause language models to misinterpret the meaning of material concepts, resulting in performance degradation in materials science tasks. Several previous studies have also shown that preserving domain-specific subwords is crucial for maintaining model effectiveness \cite{gutierrez2023biomedical, gu2021domain, hofmann2021superbizarre}, but how to identify and preserve such words remains unexplored in the materials science domain.

To address this issue, we propose \textbf{MATTER} (\textbf{Mat}erials \textbf{T}ok\textbf{e}nization F\textbf{r}amework), a novel approach that integrates material knowledge into tokenization. MATTER involves carefully designed frequency computation and merging processes to effectively capture the material concepts. We present MatDetector, a material concept identifier that scores each concept by its relevance to the materials domain, trained on a corpus of material knowledge that we carefully constructed. Subsequently, jointly considering the relevance scores and statistics of words, MATTER re-ranks the score of multiple possible merged tokens, prioritizing material-related subwords to be preserved. By integrating material knowledge into frequency computation and restructuring token merging, MATTER addresses the limitations of standard frequency-centric tokenization and enhances the representation of material concepts.

To verify the efficacy of MATTER, we conduct comprehensive experiments across diverse downstream tasks in materials science, including both generation and classification. The results demonstrate that MATTER significantly enhances performance on material-specific tasks while preserving the unique characteristics of material terminology. By integrating material knowledge into tokenization training, MATTER enables more precise learning of domain-specific concepts, underscoring the effectiveness of this tailored approach. In summary, this paper presents the following key contributions:
    \begin{itemize}
        \item We introduce MATTER, a novel domain-specific tokenization framework that integrates material knowledge into the tokenization process.
        \item We develop a novel scheme for materials tokenization based on MatDetector trained on our materials knowledge corpus integrated into our re-ranked token merging process.
        \item We demonstrate that MATTER outperforms existing tokenization methods, achieving an average improvement of 4\% on generation tasks and 2\% on classification tasks through extensive experiments.
    \end{itemize}

\section{Related Work}

\subsection{Subword Tokenization}
Tokenization plays a crucial role in the performance of language models \cite{rust2021good, singh2024tokenization, wang2024tokenization}. One significant advancement in this area is subword tokenization, a pivotal approach in NLP. Various subword tokenization techniques exist, among which frequency-centric methods, such as Byte Pair Encoding (BPE; \citealt{gage1994new,sennrich-etal-2016-neural}) and WordPiece \cite{wu2016google}, construct subword vocabularies by merging frequently co-occurring character sequences. Recent studies have explored integrating additional linguistic and contextual signals into tokenization. SAGE \cite{yehezkel2022incorporating} introduces contextual embeddings to guide token segmentation, while PickyBPE \cite{chizhov2024bpe} refines intermediate “junk” tokens.

However, while these methods effectively preserve high-frequency words, they often fragment low-frequency words, obscuring their meaning \cite{schmidt2024tokenization, wu2016google, sennrich2015neural, mikolov2012subword}. Additionally, they are designed for general-domain corpora and fail to account for specialized terminology in the materials domain, where key concepts are both semantically significant and infrequent. As a result, conventional tokenization methods frequently split material concepts into unrelated subwords, disrupting their meaning. In contrast, MATTER is designed to address these domain-specific challenges in materials science.

\begin{figure*}[t] 
    \includegraphics[width=1\textwidth]{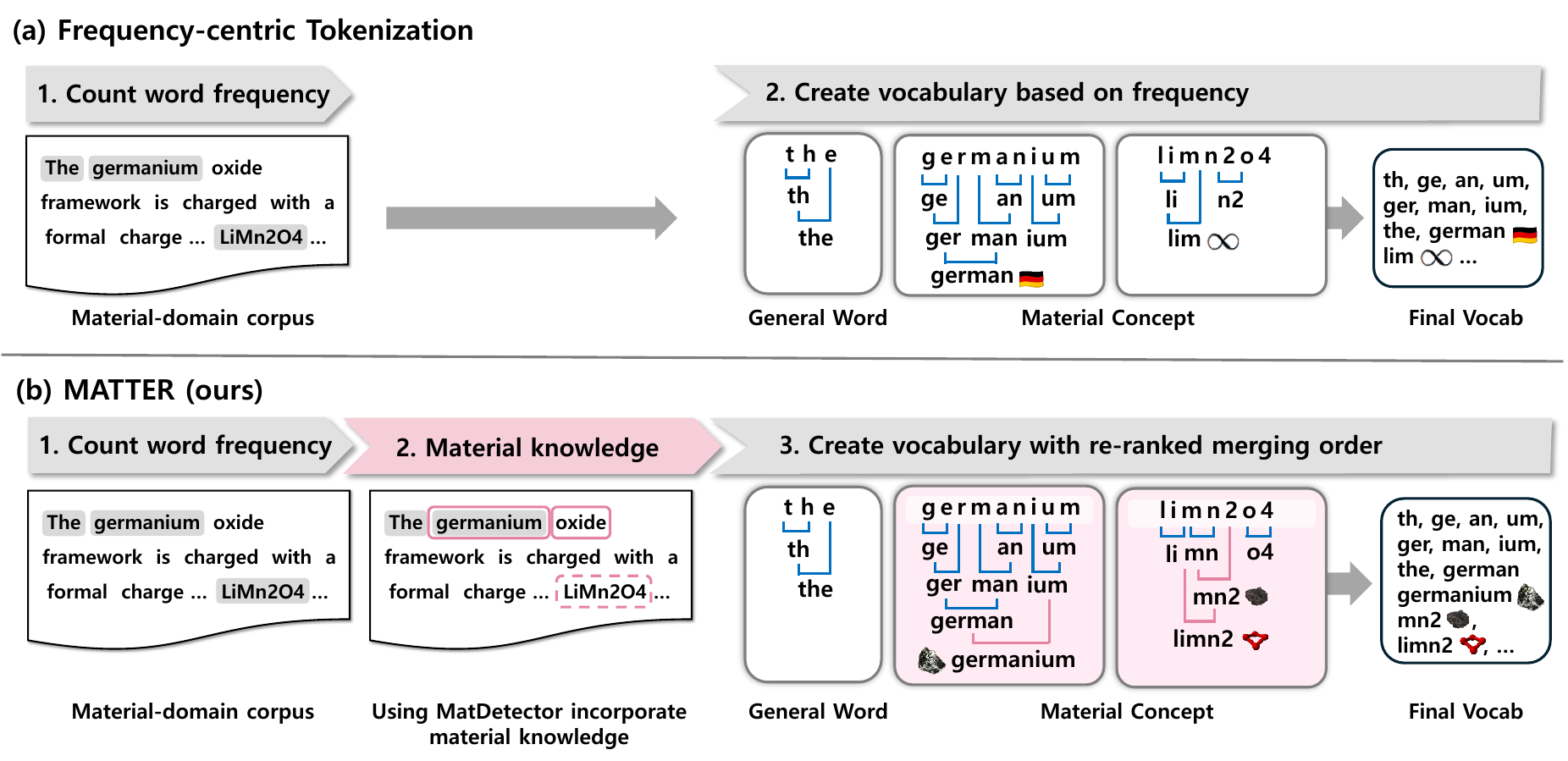} 
    \caption{Comparison of the overall methodology between the existing frequency-centric tokenization and MATTER: (a) The existing frequency-centric tokenization creates the vocabularies based on word frequency. (b) In contrast, our approach, MATTER, incorporates material knowledge from MatDetector into subword vocabularies.} 
    \label{fig:main} 
\end{figure*}

\subsection{Language Models in Materials Science}
The discovery and practical application of materials is a time-intensive process, often spanning decades \cite{national2011materials,jain2013commentary}. To accelerate this process, leveraging the wealth of knowledge captured in textual datasets has become essential. NLP-based approaches have potential in materials informatics, enabling advancements in extracting and utilizing domain-specific knowledge \cite{wang2024joint, friedrich2020sofc, weston2019named, mysore2019materials}. \citet{tshitoyan2019unsupervised} introduced embedding-based unsupervised methods, effectively capturing chemical knowledge and understanding chemical properties. Building on this foundation, \citet{trewartha2022quantifying} introduced pre-trained language models trained on a materials science corpus, utilizing BERT \cite{devlin-etal-2019-bert}. Further extending the capabilities of BERT-based models, SciBERT \cite{beltagy2019scibert}, trained on material and battery-specific corpora, was adapted into MatSciBERT \cite{gupta2022matscibert} and BatteryBERT \cite{huang2022batterybert}, respectively. 

However, these models rely on tokenization strategies originally designed for general NLP tasks, which can be suboptimal for material specialized terminology. In contrast, MATTER introduces a tokenization approach tailored to the unique linguistic characteristics of the materials domain.

\section{MATTER}
We propose a materials-aware tokenization approach that integrates material knowledge into tokenization training and re-ranks token merging order. The overall procedure is illustrated in Figure \ref{fig:main}.

\subsection{Word Frequency Calculation}
\label{sec:WordCalculation}

MATTER incorporates the WordPiece algorithm, a frequency-centric tokenization method, with material domain knowledge. The standard WordPiece algorithm first computes the frequency of each word in the corpus. Then, it tokenizes words into sequences of characters or byte units and iteratively merges the most frequent pair of tokens. Similarly, MATTER also computes word frequencies, denoted as \( \text{freq}_{\text{origin}}(w) \) for a word \( w \):

\[
\text{freq}_{\text{origin}}(w) = \text{count}(w)
\]

\noindent where \( \text{count}(w) \) represents the number of occurrences of word \( w \) in the corpus. However, MATTER further incorporates material knowledge (§ \ref{sec:knowledge}) and re-ranks the token merging order (§ \ref{sec:knowledge2}) to better preserve domain-specific terminology.

\subsection{Material Knowledge Incorporation}
\label{sec:knowledge}
To integrate material knowledge into MATTER, we adjust word frequencies (§ \ref{sec:WordCalculation}) by assigning weights to material concepts. Therefore, precise identification of material concepts is crucial. Traditionally, ChemDataExtractor \cite{kumar2024database} has been widely used in materials science for this purpose, but since it was trained on biomedical data, its accuracy in identifying material concepts is limited \cite{kim2024melt, kumar2024database, tran2024design, xu2023small}.

To address this, we introduce MatDetector, a material-agnostic tool that detects material concepts in a target corpus and assigns probability scores to each concept. Developed using the architecture of \citet{trewartha2022quantifying}, MatDetector is optimized for material concept detection. The dataset creation process is as follows:

 \paragraph{Material Concept Extraction} The MatDetector searches the PubChem database \cite{kim2019pubchem} for material-related concepts, extracting 80K material concepts (chemical names, IUPAC names, synonyms, and molecular formulas).

 \paragraph{Material Corpus Crawling} Using these concepts extracted from PubChem, we crawl Semantic Scholar, collecting around 42K scientific papers.

 \paragraph{Crawled Data Tagging} The collected corpus is tagged with PubChem material concepts, creating a NER material dataset with labels "material name", "material formula", and "other". Labels other than "other" are treated as "material concept".

 \paragraph{Data Augmentation} While Semantic Scholar offers relatively clean data, material-related datasets from journals and repositories often contain formatting inconsistencies, OCR errors, and structural variations. To address this, we standardized common noise and expanded the dataset fourfold to enhance model robustness. Details in Appendix \ref{sec:MatDetector}.

\begin{algorithm}[t] 
\caption{MATTER Tokenization Training}
\label{alg:matter_tokenization}
{\small 
\textbf{Input:} Corpus $C$, Vocabulary size $V$, MatDetector $MD$, Material importance factor $\lambda$ \\
\textbf{Output:} Vocabulary $\mathcal{V}$ of size $V$ (ordered)
\begin{algorithmic}[1]
\Procedure{MATTER}{$C, V, MD, \lambda$}
    \State $\mathcal{V} \leftarrow \{ c \mid c \in C \}$ \Comment{Unique characters}
    \State $\text{freq}_{\text{origin}}(w) \leftarrow {\text{word frequency for all }C}$
    \State $\hat{y}_\text{mat}(w) \leftarrow MD(w)$
    \ForAll{$w \in C$} \Comment{Re-ranking}
        \If{$\hat{y}_\text{mat}(w) \neq \emptyset$} 
            \State $\text{freq}_{\text{mat}}(w) \gets \text{freq}_{\text{origin}}(w) + \lambda \cdot \frac{\hat{y}_\text{mat}(w)}{1 - \hat{y}_\text{mat}(w)}$
        \Else 
            \State $\text{freq}_{\text{mat}}(w) \gets \text{freq}_{\text{origin}}(w)$
        \EndIf
    \EndFor
    
    \State Compute $\text{Score}(t_L, t_R)$ for all token pairs
    
    \While{$|\mathcal{V}| < V$} \Comment{Merge tokens}
        \State $\langle t_L, t_R \rangle \gets \arg\max_{(t_L, t_R)} \text{MatScore}(t_L, t_R)$
        \State $t_{\text{new}} \gets t_L \oplus t_R$ \Comment{Create new token}
        \State $\mathcal{V} \gets \mathcal{V} \cup \{t_{\text{new}}\}$ 
        \State $C.\text{ReplaceAll}(\langle t_L, t_R \rangle, t_{\text{new}})$ \Comment{Update corpus}
        \State Recompute $\text{Score}$ for updated token set
    \EndWhile

    \State \Return $\mathcal{V}$
\EndProcedure
\end{algorithmic}
} 
\end{algorithm}

Using the MatDetector, we can detect material concepts and compute their probability. Specifically, for Given a word \( w \) that is split into \( n \) subword tokens \( \{t_1, t_2, ..., t_n\} \), the label for the word is determined as follows:
\begin{equation} 
    \label{eq:1}
    \hat{y}(w) = \arg\max_{c \in C} \frac{1}{n} \sum_{i=1}^{n} P(t_i, c)
\end{equation}

\noindent where \( C \) is the set of all possible labels, and \( P(t_i, c) \) denotes the probability of subword token \( t_i \) being classified as label \( c \). If the predicted label \( \hat{y}(w) \) falls under "material concept," we denote it as \( \hat{y}_\text{mat}(w) \). The equation is as follows:

\begin{equation} 
    \label{eq:2}
    \hat{y}_\text{mat}(w) =
    \begin{cases} 
        \hat{y}(w), & \text{if } \hat{y}(w) \in \{\text{material}\} \\
        \emptyset, & \text{otherwise}
    \end{cases}
\end{equation}

Ultimately, material concepts identified within the vocabulary are assigned \( \hat{y}_\text{mat}(w) \), representing the likelihood of a word being relevant to the material domain. A higher probability value indicates stronger relevance to material concepts, ensuring that domain-specific concepts are effectively distinguished from general words.

\subsection{Vocab Creation with Re-ranked Order} 
\label{sec:knowledge2}

To integrate \( \hat{y}_\text{mat}(w) \) into tokenization, we adjust word frequency computations by weighting material concepts based on their assigned probability scores. This adjustment prevents material concepts from being underrepresented, preserving their structural and semantic integrity during tokenization. To incorporate material information, we assign weighted frequencies to material concepts as follows:
 
 Using this \( \hat{y}_\text{mat}(w) \), MATTER adjusts the original frequency to prioritize material concepts. The adjusted frequency is computed as follows:

\begin{equation}
    \label{eq:example}
    \text{freq}_{\text{mat}}(w) = \text{freq}_{\text{origin}}(w) + \lambda \cdot \frac{\hat{y}_\text{mat}(w) }{1 - \hat{y}_\text{mat}(w) }
\end{equation}

With the adjusted frequency incorporating material knowledge, MATTER re-ranks the merging order based on incorporated material knowledge. Words are initially decomposed into sequences of characters or byte units, and the algorithm iteratively merges token pairs according to the re-ranked order guided by material relevance. The detailed algorithm is provided in Algorithm \ref{alg:matter_tokenization}.

\begin{table*}[ht]
\centering
\arraybackslash
\fontsize{8}{10.2}\selectfont 
\setlength{\tabcolsep}{3pt} 
\renewcommand{\arraystretch}{1.3} 
\resizebox{\textwidth}{!}{%
\begin{tabular}{cccccccccc}
\toprule
 &  & \multicolumn{8}{c}{Generation Task} \\ \cmidrule{3-10} 
\multirow{-2}{*}{Tokenization} & \multirow{-2}{*}{Metric} & NER & RC & EAE & PC & SAR & SC & SF & Overall \\ \midrule 
 & Micro-F1 & ${55.7}_\text{\textcolor{gray}{±0.4}}$ & ${49.3}_\text{\textcolor{gray}{±0.2}}$ & ${48.3}_\text{\textcolor{gray}{±0.8}}$ & ${67.3}_\text{\textcolor{gray}{±0.1}}$ & ${61.1}_\text{\textcolor{gray}{±1.8}}$ & ${90.7}_\text{\textcolor{gray}{±2.4}}$ & ${36.3}_\text{\textcolor{gray}{±1.4}}$ & ${63.5}_\text{\textcolor{gray}{±0.5}}$ \\
\multirow{-2}{*}{\begin{tabular}[c]{@{}c@{}}BPE   \\ \cite{sennrich-etal-2016-neural}\end{tabular}} & Macro-F1 & ${47.1}_\text{\textcolor{gray}{±0.5}}$ & ${47.2}_\text{\textcolor{gray}{±0.9}}$ & ${36.3}_\text{\textcolor{gray}{±0.3}}$ & ${40.2}_\text{\textcolor{gray}{±0.0}}$ & ${41.8}_\text{\textcolor{gray}{±1.3}}$ & ${47.6}_\text{\textcolor{gray}{±0.0}}$ & ${16.7}_\text{\textcolor{gray}{±1.6}}$ & ${42.0}_\text{\textcolor{gray}{±0.9}}$ \\ \midrule 

 & Micro-F1 & ${76.6}_\text{\textcolor{gray}{±0.2}}$ & ${80.9}_\text{\textcolor{gray}{±0.3}}$ & ${48.5}_\text{\textcolor{gray}{±0.2}}$ & ${73.1}_\text{\textcolor{gray}{±0.5}}$ & ${81.9}_\text{\textcolor{gray}{±0.4}}$ & ${90.0}_\text{\textcolor{gray}{±0.1}}$ & ${57.4}_\text{\textcolor{gray}{±0.2}}$ & ${72.6}_\text{\textcolor{gray}{±0.1}}$ \\
\multirow{-2}{*}{\begin{tabular}[c]{@{}c@{}}WordPiece   \\ \cite{wu2016google}\end{tabular}} & Macro-F1 & ${56.1}_\text{\textcolor{gray}{±0.2}}$ & ${58.5}_\text{\textcolor{gray}{±0.6}}$ & ${29.4}_\text{\textcolor{gray}{±0.3}}$ & ${58.9}_\text{\textcolor{gray}{±1.0}}$ & ${74.6}_\text{\textcolor{gray}{±0.9}}$ & ${60.3}_\text{\textcolor{gray}{±0.8}}$ & ${32.6}_\text{\textcolor{gray}{±0.2}}$ & ${52.9}_\text{\textcolor{gray}{±0.2}}$ \\ \midrule 
 & Micro-F1 & ${77.0}_\text{\textcolor{gray}{±0.2}}$ & ${82.3}_\text{\textcolor{gray}{±0.4}}$ & ${47.3}_\text{\textcolor{gray}{±0.1}}$ & ${68.3}_\text{\textcolor{gray}{±0.8}}$ & ${77.1}_\text{\textcolor{gray}{±0.4}}$ & ${90.9}_\text{\textcolor{gray}{±0.1}}$ & ${57.1}_\text{\textcolor{gray}{±0.3}}$ & ${71.4}_\text{\textcolor{gray}{±0.2}}$ \\
\multirow{-2}{*}{\begin{tabular}[c]{@{}c@{}}SAGE   \\ \cite{yehezkel2022incorporating}\end{tabular}} & Macro-F1 & ${57.0}_\text{\textcolor{gray}{±0.3}}$ & ${61.6}_\text{\textcolor{gray}{±0.4}}$ & ${28.3}_\text{\textcolor{gray}{±0.3}}$ & ${59.6}_\text{\textcolor{gray}{±1.3}}$ & ${67.4}_\text{\textcolor{gray}{±0.9}}$ & ${61.6}_\text{\textcolor{gray}{±0.8}}$ & ${35.0}_\text{\textcolor{gray}{±0.3}}$ & ${52.9}_\text{\textcolor{gray}{±0.3}}$ \\ \midrule 
 & Micro-F1 & ${55.4}_\text{\textcolor{gray}{±0.1}}$ & $\textbf{92.1}_\text{\textcolor{gray}{±0.1}}$ & ${47.9}_\text{\textcolor{gray}{±0.4}}$ & ${67.2}_\text{\textcolor{gray}{±0.0}}$ & ${75.7}_\text{\textcolor{gray}{±0.2}}$ & ${90.7}_\text{\textcolor{gray}{±0.0}}$ & ${43.6}_\text{\textcolor{gray}{±0.1}}$ & ${67.5}_\text{\textcolor{gray}{±0.1}}$ \\
\multirow{-2}{*}{\begin{tabular}[c]{@{}c@{}}PickyBPE  \\ \cite{chizhov2024bpe}\end{tabular}} & Macro-F1 & ${41.7}_\text{\textcolor{gray}{±0.1}}$ & $\textbf{65.1}_\text{\textcolor{gray}{±0.2}}$ & ${36.5}_\text{\textcolor{gray}{±0.6}}$ & ${40.2}_\text{\textcolor{gray}{±0.0}}$ & ${66.1}_\text{\textcolor{gray}{±0.7}}$ & ${47.6}_\text{\textcolor{gray}{±0.0}}$ & ${23.1}_\text{\textcolor{gray}{±0.1}}$ & ${45.8}_\text{\textcolor{gray}{±0.1}}$ \\ \midrule 
\rowcolor[HTML]{E9EBF5} 
\cellcolor[HTML]{E9EBF5} & Micro-F1 & $\textbf{80.0}_\text{\textcolor{gray}{±0.0}}$ & ${83.8}_\text{\textcolor{gray}{±0.1}}$ & $\textbf{53.1}_\text{\textcolor{gray}{±0.2}}$ & $\textbf{73.7}_\text{\textcolor{gray}{±0.2}}$ & $\textbf{85.5}_\text{\textcolor{gray}{±0.3}}$ & $\textbf{91.2}_\text{\textcolor{gray}{±0.1}}$ & $\textbf{61.9}_\text{\textcolor{gray}{±0.3}}$ & $\textbf{75.6}_\text{\textcolor{gray}{±0.1}}$ \\
\rowcolor[HTML]{E9EBF5} 
\multirow{-2}{*}{\cellcolor[HTML]{E9EBF5}MATTER   (ours)} & Macro-F1 & $\textbf{59.3}_\text{\textcolor{gray}{±0.2}}$ & ${59.1}_\text{\textcolor{gray}{±0.5}}$ & $\textbf{36.9}_\text{\textcolor{gray}{±0.3}}$ & $\textbf{67.6}_\text{\textcolor{gray}{±0.6}}$ & $\textbf{79.3}_\text{\textcolor{gray}{±0.7}}$ & $\textbf{64.9}_\text{\textcolor{gray}{±0.5}}$ & $\textbf{38.0}_\text{\textcolor{gray}{±0.3}}$ & $\textbf{57.9}_\text{\textcolor{gray}{±0.1}}$ \\ \bottomrule

\end{tabular}

}

  \caption{Evaluation results on MatSci-NLP (generation tasks): The tasks encompass Named Entity Recognition (NER), Relation Classification (RC), Event Argument Extraction (EAE), Paragraph Classification (PC), Synthesis Action Retrieval (SAR), Sentence Classification (SC), and Slot Filling (SF). The best-performing results are highlighted in \textbf{boldface}.}
  \label{tab:accents}

\end{table*}

\begin{table*}[ht]
\centering
\arraybackslash
\fontsize{8}{10.2}\selectfont 
\setlength{\tabcolsep}{3pt} 
\renewcommand{\arraystretch}{1.3} 
\resizebox{\textwidth}{!}{%
\begin{tabular}{cccccccccccc}
\toprule 
\cellcolor[HTML]{FFFFFF} & \cellcolor[HTML]{FFFFFF} & \multicolumn{10}{c}{\cellcolor[HTML]{FFFFFF}Classification Task} \\ \cmidrule{3-12} 
\cellcolor[HTML]{FFFFFF} & \cellcolor[HTML]{FFFFFF} & \multicolumn{2}{c}{\cellcolor[HTML]{FFFFFF} NER$_\text{SOFC}$} & \multicolumn{2}{c}{\cellcolor[HTML]{FFFFFF}NER$_\text{Matscholar}$} & \multicolumn{2}{c}{\cellcolor[HTML]{FFFFFF}SF} & \multicolumn{2}{c}{\cellcolor[HTML]{FFFFFF}RC} & \multicolumn{2}{c}{\cellcolor[HTML]{FFFFFF}PC*} \\  
\multirow{-3}{*}{\cellcolor[HTML]{FFFFFF}Tokenization} & \multirow{-3}{*}{\cellcolor[HTML]{FFFFFF}Metric} & val & test & val & test & val & test & val & test & val & test \\ \midrule 
\rowcolor[HTML]{FFFFFF} 
\cellcolor[HTML]{FFFFFF} & Micro-F1 & ${81.6}_\text{\textcolor{gray}{±0.2}}$ & ${81.4}_\text{\textcolor{gray}{±0.1}}$ & ${86.4}_\text{\textcolor{gray}{±0.3}}$ & ${84.3}_\text{\textcolor{gray}{±0.5}}$ & ${68.1}_\text{\textcolor{gray}{±0.5}}$ & ${68.3}_\text{\textcolor{gray}{±0.6}}$ & ${90.2}_\text{\textcolor{gray}{±0.4}}$ & ${89.9}_\text{\textcolor{gray}{±0.0}}$ & \cellcolor[HTML]{FFFFFF} & \cellcolor[HTML]{FFFFFF} \\
\rowcolor[HTML]{FFFFFF} 
\multirow{-2}{*}{\cellcolor[HTML]{FFFFFF}\begin{tabular}[c]{@{}c@{}}BPE\\      \cite{sennrich-etal-2016-neural}\end{tabular}} & Macro-F1 & ${80.7}_\text{\textcolor{gray}{±0.2}}$ & ${78.9}_\text{\textcolor{gray}{±0.1}}$ & ${85.0}_\text{\textcolor{gray}{±0.6}}$ & ${82.9}_\text{\textcolor{gray}{±0.7}}$ & ${65.5}_\text{\textcolor{gray}{±0.4}}$ & $\text{59.3}_\text{\textcolor{gray}{±0.8}}$ & ${86.4}_\text{\textcolor{gray}{±0.1}}$ & ${85.5}_\text{\textcolor{gray}{±0.1}}$ & \multirow{-2}{*}{\cellcolor[HTML]{FFFFFF}${95.5}_\text{\textcolor{gray}{±0.0}}$} & \multirow{-2}{*}{\cellcolor[HTML]{FFFFFF}${95.6}_\text{\textcolor{gray}{±0.0}}$} \\ \midrule

\rowcolor[HTML]{FFFFFF} 
\cellcolor[HTML]{FFFFFF} & Micro-F1 & ${82.0}_\text{\textcolor{gray}{±0.6}}$ & ${80.9}_\text{\textcolor{gray}{±0.4}}$ & ${88.8}_\text{\textcolor{gray}{±0.2}}$ & ${86.1}_\text{\textcolor{gray}{±0.3}}$ & ${67.4}_\text{\textcolor{gray}{±0.5}}$ & $\textbf{60.4}_\text{\textcolor{gray}{±0.7}}$ & ${90.6}_\text{\textcolor{gray}{±0.2}}$ & ${91.0}_\text{\textcolor{gray}{±0.7}}$ & \cellcolor[HTML]{FFFFFF} & \cellcolor[HTML]{FFFFFF} \\
\rowcolor[HTML]{FFFFFF} 
\multirow{-2}{*}{\cellcolor[HTML]{FFFFFF}\begin{tabular}[c]{@{}c@{}}WordPiece\\      \cite{wu2016google}\end{tabular}} & Macro-F1 & ${83.0}_\text{\textcolor{gray}{±0.2}}$ & ${83.0}_\text{\textcolor{gray}{±0.4}}$ & ${87.6}_\text{\textcolor{gray}{±0.3}}$ & ${85.8}_\text{\textcolor{gray}{±0.2}}$ & ${69.2}_\text{\textcolor{gray}{±0.4}}$ & ${69.6}_\text{\textcolor{gray}{±0.4}}$ & ${86.3}_\text{\textcolor{gray}{±0.3}}$ & ${87.5}_\text{\textcolor{gray}{±0.1}}$ & \multirow{-2}{*}{\cellcolor[HTML]{FFFFFF}${95.2}_\text{\textcolor{gray}{±0.1}}$} & \multirow{-2}{*}{\cellcolor[HTML]{FFFFFF}${95.2}_\text{\textcolor{gray}{±0.1}}$} \\ \midrule 

\rowcolor[HTML]{FFFFFF} 
\cellcolor[HTML]{FFFFFF} & Micro-F1 & ${82.0}_\text{\textcolor{gray}{±0.2}}$ & ${79.7}_\text{\textcolor{gray}{±0.4}}$ & ${88.4}_\text{\textcolor{gray}{±0.3}}$ & ${86.7}_\text{\textcolor{gray}{±0.4}}$ & ${67.9}_\text{\textcolor{gray}{±0.5}}$ & ${60.3}_\text{\textcolor{gray}{±0.4}}$ & ${89.8}_\text{\textcolor{gray}{±0.4}}$ & ${90.6}_\text{\textcolor{gray}{±0.3}}$ &  \cellcolor[HTML]{FFFFFF} & \cellcolor[HTML]{FFFFFF} \\
\rowcolor[HTML]{FFFFFF} 
\multirow{-2}{*}{\cellcolor[HTML]{FFFFFF}\begin{tabular}[c]{@{}c@{}}SAGE\\      \cite{yehezkel2022incorporating}\end{tabular}} & Macro-F1 & ${82.7}_\text{\textcolor{gray}{±0.2}}$ & ${82.5}_\text{\textcolor{gray}{±0.8}}$ & ${87.6}_\text{\textcolor{gray}{±0.2}}$ & ${86.1}_\text{\textcolor{gray}{±0.1}}$ & $\textbf{69.7}_\text{\textcolor{gray}{±0.3}}$ & ${69.5}_\text{\textcolor{gray}{±0.6}}$ & ${86.4}_\text{\textcolor{gray}{±0.7}}$ & ${87.1}_\text{\textcolor{gray}{±0.0}}$ & \multirow{-2}{*}{\cellcolor[HTML]{FFFFFF}${95.3}_\text{\textcolor{gray}{±0.0}}$} & \multirow{-2}{*}{\cellcolor[HTML]{FFFFFF}${95.6}_\text{\textcolor{gray}{±0.2}}$} \\ \midrule

\rowcolor[HTML]{FFFFFF} 
\cellcolor[HTML]{FFFFFF} & Micro-F1 & ${77.3}_\text{\textcolor{gray}{±0.3}}$ & ${78.8}_\text{\textcolor{gray}{±0.6}}$ & ${84.1}_\text{\textcolor{gray}{±0.4}}$ & ${83.4}_\text{\textcolor{gray}{±0.6}}$ & ${62.0}_\text{\textcolor{gray}{±0.3}}$ & ${60.2}_\text{\textcolor{gray}{±0.4}}$ & ${88.6}_\text{\textcolor{gray}{±0.1}}$ & ${85.8}_\text{\textcolor{gray}{±0.2}}$ &  \cellcolor[HTML]{FFFFFF} & \cellcolor[HTML]{FFFFFF} \\
\rowcolor[HTML]{FFFFFF} 
\multirow{-2}{*}{\cellcolor[HTML]{FFFFFF}\begin{tabular}[c]{@{}c@{}}PickyBPE\\      \cite{chizhov2024bpe}\end{tabular}} & Macro-F1 & ${78.6}_\text{\textcolor{gray}{±0.4}}$ & ${81.0}_\text{\textcolor{gray}{±0.7}}$ & ${86.1}_\text{\textcolor{gray}{±0.3}}$ & ${84.7}_\text{\textcolor{gray}{±0.5}}$ & ${67.1}_\text{\textcolor{gray}{±0.1}}$ & ${55.4}_\text{\textcolor{gray}{±0.2}}$ & ${88.8}_\text{\textcolor{gray}{±0.6}}$ & ${87.0}_\text{\textcolor{gray}{±0.2}}$ & \multirow{-2}{*}{\cellcolor[HTML]{FFFFFF}${95.7}_\text{\textcolor{gray}{±0.3}}$} & \multirow{-2}{*}{\cellcolor[HTML]{FFFFFF}${95.8}_\text{\textcolor{gray}{±0.2}}$} \\ \midrule

\rowcolor[HTML]{E9EBF5} 
\cellcolor[HTML]{E9EBF5} & Micro-F1 & $\textbf{83.1}_\text{\textcolor{gray}{±0.2}}$ & $\textbf{82.0}_\text{\textcolor{gray}{±0.4}}$ & $\textbf{89.6}_\text{\textcolor{gray}{±0.1}}$ & $\textbf{87.8}_\text{\textcolor{gray}{±0.4}}$ & $\textbf{68.4}_\text{\textcolor{gray}{±0.1}}$ & $\textbf{60.4}_\text{\textcolor{gray}{±0.4}}$ & $\textbf{90.9}_\text{\textcolor{gray}{±0.2}}$ & $\textbf{92.6}_\text{\textcolor{gray}{±0.6}}$ &\cellcolor[HTML]{E9EBF5} & \cellcolor[HTML]{E9EBF5} \\
\rowcolor[HTML]{E9EBF5} 
\multirow{-2}{*}{\cellcolor[HTML]{E9EBF5}MATTER   (ours)} & Macro-F1  & $\textbf{84.3}_\text{\textcolor{gray}{±0.2}}$ & $\textbf{84.4}_\text{\textcolor{gray}{±0.3}}$ & $\textbf{88.6}_\text{\textcolor{gray}{±0.2}}$ & $\textbf{86.3}_\text{\textcolor{gray}{±0.3}}$ & $\textbf{69.7}_\text{\textcolor{gray}{±0.4}}$ & $\textbf{70.1}_\text{\textcolor{gray}{±0.3}}$ & $\textbf{87.3}_\text{\textcolor{gray}{±0.4}}$ & $\textbf{87.9}_\text{\textcolor{gray}{±0.9}}$ & \multirow{-2}{*}{\cellcolor[HTML]{E9EBF5}$\textbf{96.9}_\text{\textcolor{gray}{±0.1}}$} & \multirow{-2}{*}{\cellcolor[HTML]{E9EBF5}$\textbf{96.2}_\text{\textcolor{gray}{±0.2}}$} \\ \bottomrule

\end{tabular}
}

  \caption{Evaluation results are presented across five classification tasks. Here, PC* represents accuracy, while the remaining metrics are reported as Micro-F1 and Macro-F1 scores. The best-performing results are highlighted in \textbf{boldface}.}
  \label{tab:accents2}
\end{table*}

\begin{table}[t] 
\centering
\resizebox{0.5\textwidth}{!}{ 
\begin{tabular}{lcccc}
\toprule
                & {Train}   & {Dev}    & {Test}   & {Total}   \\
\midrule
English Set     & 458{,}692 & 57{,}371 & 57{,}755 & 573{,}818 \\
Material Subset & 16{,}286  & 2{,}010  & 2{,}173  & 20{,}469 \\
\bottomrule
\end{tabular}
}
\caption{
Statistics for the SIGMORPHON 2022 morpheme segmentation dataset and the material dataset, as described in Section~\ref{sec:material_morpheme}.
}
\label{tab:set}
\end{table}

\renewcommand{\arraystretch}{1.1}  
\begin{table}[t]
\centering
\resizebox{0.5\textwidth}{!}{ 
\begin{tabular}{lcc} 
\toprule
Tokenization for MatSciBERT & Segmentation F1  \\ 
\midrule
WordPiece \cite{wu2016google}                               & 44.3                      \\
SAGE \cite{yehezkel2022incorporating}                                    & 43.4                      \\
PickyBPE \cite{chizhov2024bpe}                               & 36.2                      \\ 
\midrule \midrule 
MATTER (ours)                              & \textbf{59.9}             \\
\bottomrule
\end{tabular}
}
\caption{
Material morpheme segmentation performance for different tokenization of the MatSciBERT model. The best-performing results are highlighted in \textbf{boldface}.}
\label{tab:segmentation_f1}
\end{table}

\begin{table}[t]
\centering
{\footnotesize
\begin{tabular}{l|ccc}
\toprule
Tool & Recall & Precision & F1 Score \\ \midrule 
ChemDataExtractor & 18\% & 57\% & 27\% \\
MatDetector (ours) & \textbf{57\%} & \textbf{69\%} & \textbf{63\%} \\ \bottomrule
\end{tabular}
}
\caption{Average performance of two material concept extraction tools on external materials NER datasets across all evaluation metrics.}
\label{tab:nerner}
\end{table}

\section{Experiments}
\subsection{Experimental Setups} 
\paragraph{Baselines} 
To verify the efficacy of MATTER, we mainly compare ours with strong tokenization method: BPE \cite{sennrich-etal-2016-neural}, WordPiece \cite{wu2016google}, SAGE \cite{yehezkel2022incorporating}, PickyBPE \cite{chizhov2024bpe}. More hyperparameters are detailed
in Appendix~\ref{sec:Implementationb1}. The detailed experimental setups are described in Appendix~\ref{sec:Implementationb2}.

\paragraph{Pre-training}  
To evaluate the impact of tokenization on performance, we trained models using both baseline and MATTER specifically for the domain of materials science. Consistent with prior methodology~\cite{gupta2022matscibert}, we adopt SciBERT~\cite{beltagy2019scibert} as the encoder backbone for all experiments, due to its widespread use in materials-specific language modeling~\cite{gupta2022matscibert, huang2022batterybert, kim2024melt}.  
All models are trained with a fixed vocabulary size of 31,090 and a corpus of 150K materials science papers. All training conditions—including model architecture, optimizer, and learning rate—are held constant across tokenizers to ensure fair comparison.  
In MATTER, the weighting parameter \( \lambda \) was set to 1 based on empirical analysis (see §~\ref{sec:ab}), and further implementation details are provided in Appendix~\ref{sec:Implementationb2}.

\paragraph{Downstream Tasks and Datasets} 
To comprehensively evaluate the performance of MATTER, we compare models trained with different tokenization methods on both generation and classification tasks. For generation tasks, we assess each baseline on the MatSci-NLP dataset \cite{song2023matsci}, which includes seven materials-related tasks. We follow the MatSci-NLP benchmark protocol, which evaluates domain-specific encoders using a transformer-based schema decoder tailored for generation-based tasks. For classification tasks, we adopt four distinct benchmarks from prior work \cite{gupta2022matscibert}, including named entity recognition \cite{weston2019named, friedrich2020sofc}, paragraph classification \cite{venugopal2021looking}, and slot filling \cite{friedrich2020sofc}. These classification models are evaluated under standard encoder-only settings as used in prior work \cite{gupta2022matscibert}. Detailed descriptions of evaluation metrics are provided in Appendix~\ref{sec:Implementationbem}.

\subsection{Main Results}

\paragraph{Generation Tasks} Table~\ref{tab:accents} shows that MATTER outperforms existing tokenization methods, boosting Micro-F1 and Macro-F1 by 3\% and 5\% on average. These gains highlight MATTER's broad applicability across materials science tasks. Notably, SAGE and PickyBPE, which introduce non-material-specific signals, perform worse than WordPiece, emphasizing the need for domain-specific knowledge in tokenization. To further examine the generalizability of MATTER, we additionally evaluate its performance on materials-domain QA tasks using decoder-based and encoder-decoder models.\footnote{See Appendix~\ref{sec:qa-exp} for full details and results. MATTER outperforms other tokenization methods on the MaScQA benchmark, showing consistent gains in both model types.}

\paragraph{Classification Tasks} Similar to the generation tasks (Table~\ref{tab:accents2}), classification results confirm MATTER's superiority, with an average Micro-F1 and Macro-F1 improvement of 1.6\% and 1.8\%, respectively. These consistent gains highlight its robustness and ability to generalize across diverse materials science contexts, reinforcing its impact on materials informatics.  To rigorously verify that these improvements are not attributed to random variation, we conducted paired t-tests for both generation and classification tasks. The detailed statistical analysis is presented in Appendix~\ref{sec:Statistical_Significance}, confirming that MATTER’s performance gains are statistically significant across all major benchmarks.

\subsection{Material Morpheme Segmentation} 
\label{sec:material_morpheme}
To validate MATTER's ability to segment material concepts into meaningful subwords, we evaluated its performance on the material subset of the SIGMORPHON dataset \cite{batsuren2022sigmorphon}. The SIGMORPHON 2022 Shared Task provides a reliable benchmark for assessing whether words are segmented into morphologically meaningful units. For this analysis, we identified material concepts shared between SIGMORPHON, PubChem, and MatKG \cite{venugopal2022largest}. The resulting subset, as shown in Table \ref{tab:set}, revealed that approximately 20\% of annotated words are relevant material concepts.

Using this subset, we evaluated the morpheme segmentation. As shown in Table \ref{tab:segmentation_f1}, MATTER achieved an average improvement of 18.6\% in segmentation accuracy compared to other tokenization algorithms. These results confirm that MATTER tokenization, effectively incorporates the characteristics of material corpora, enabling it to segment material concepts into meaningful subwords.

\subsection{Extracted Material Concepts} 
\paragraph{Validation on Training Corpus} To validate MatDetector on the training corpus, we constructed a reference lexicon of ~100K material-related entries from PubChem and MatKG, including names, formulas, and synonyms. These were decomposed into ~1.6M normalized tokens for broader coverage. Entities extracted from 150K materials papers were matched to the lexicon, and considered valid if found in the lexicon. MatDetector extracted 6× more material concepts than ChemDataExtractor and achieved 64\% higher match rate, confirming its precision and suitability for identifying material concepts in materials science corpus.

\paragraph{Validation on Materials NER} To quantify absolute performance, we additionally evaluate both tools on two external materials NER datasets: MatScholar~\cite{weston2019named} and SOFC~\cite{friedrich2020sofc}. As shown in Table~\ref{tab:nerner}, which reports the average performance across the two datasets, MatDetector consistently outperforms ChemDataExtractor across precision, recall, and F1 score. Notably, it achieves over twice the F1 score on average, highlighting its effectiveness not only in coverage but also in accurately identifying material entities. Detailed per-dataset results are provided in Appendix \ref{sec:NER}. These results further validate MatDetector’s ability to accurately and comprehensively detect material concepts in domain-specific NER tasks.

\subsection{Token Qualities} 
To assess material token quality, we extract material-related tokens using MatDetector and compare tokenization methods.More hyperparameters are detailed in Appendix~\ref{sec:qualieites_setting}.

\begin{figure}[t] 
    \centering 

    \includegraphics[width=0.48\textwidth]{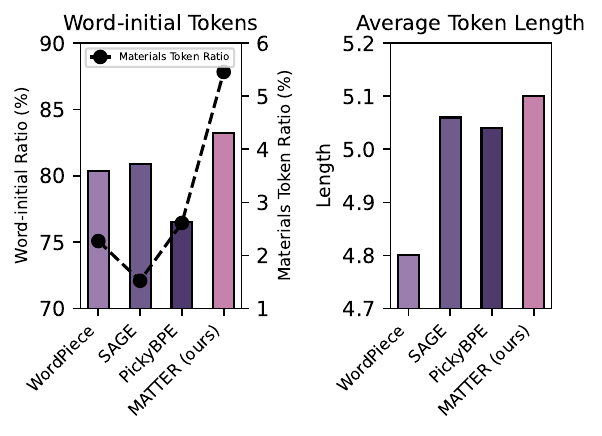} 
    \caption{Comparison tokenization methods by word-initial token ratio (bar), materials token ratio (line), and average token length.} 
    \label{fig:Token_Q} 
\end{figure}

\paragraph{Word-Initial Token} One key aspect of token quality is the proportion of word-initial tokens, which help preserve word structure and meaning \cite{yehezkel2022incorporating, chizhov2024bpe}. For example, in tokenizing \textit{"germanium"} into \textit{"german"} and \textit{"-ium"}, the word-initial token is "german". As shown in the left part of Figure~\ref{fig:Token_Q}, MATTER preserves a higher proportion of word-initial tokens (bar) compared to other methods. To evaluate this more rigorously, we further measured the materials-related word-initial token ratio (line) using a manually annotated set of approximately 9,000 material concepts, curated for downstream evaluation only (details in Appendix~\ref{sec:Details_Word-Initial_Token}). While this represents a small fraction of the full corpus, the results consistently demonstrate that MATTER achieves a significantly higher proportion of materials-related word-initial tokens, even on unseen datasets. This indicates that its vocabulary is enriched with material-specific terms, enabling better preservation of the semantic integrity of materials-related concepts.

 \paragraph{Token Length}
According to \citet{bostrom2020byte}, longer mean token length reflects gold-standard morphologically-aligned tokenization, which enhances token quality. Based on this, we also measure mean token length. As shown in the right part of Figure \ref{fig:Token_Q}, our method achieves a higher mean token length. Notably, it surpasses even SAGE and PickyBPE, which deliberately eliminate shorter intermediate tokens through compression at the cost of increased computational expense. This demonstrates that our approach not only maintains morphological alignment for material concepts but also preserves higher-quality tokenization.

\begin{figure}[t] 
    \centering 
    \includegraphics[width=0.48\textwidth]{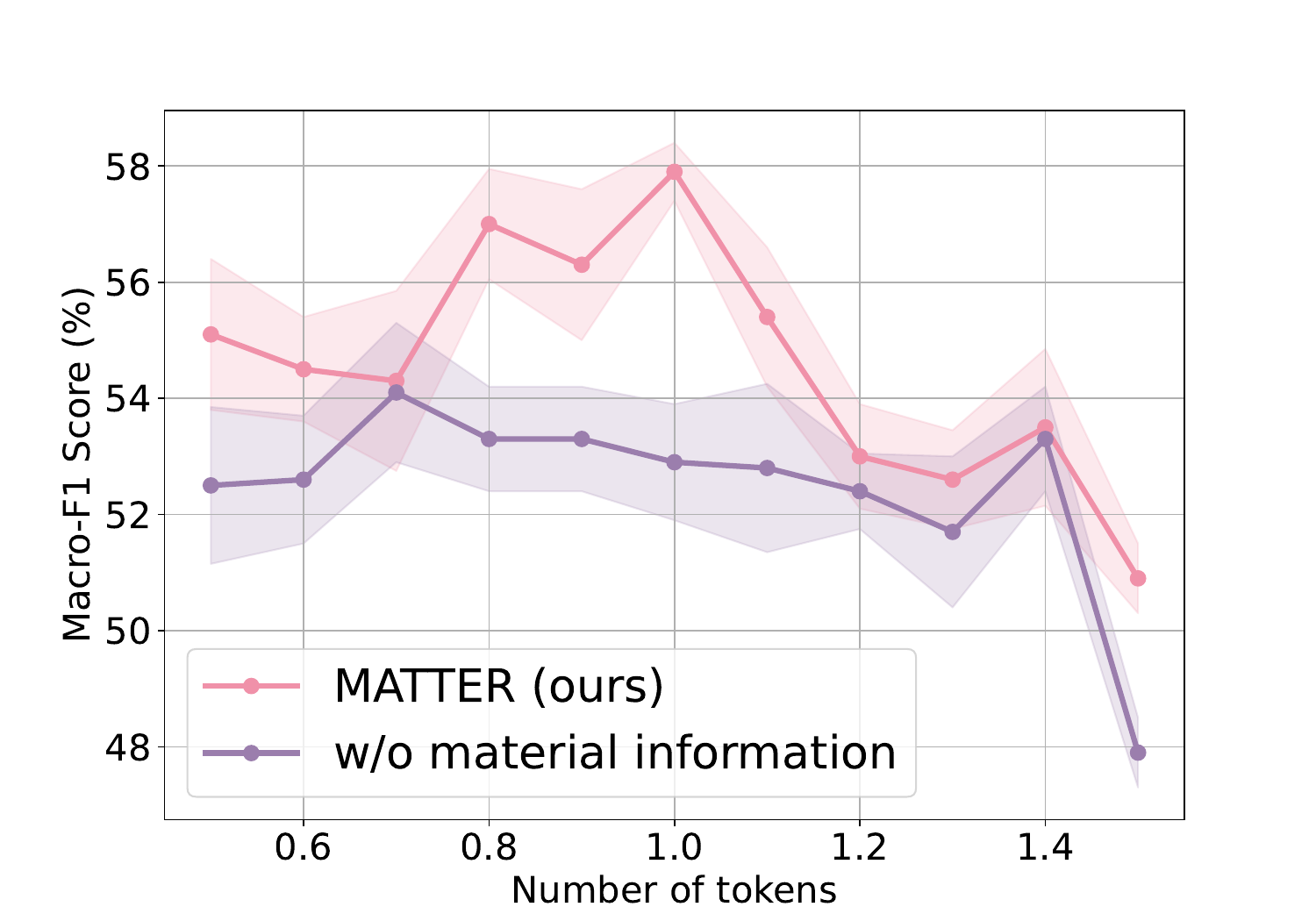} 

    \caption{Comparison of Macro-F1 scores for MATTER and w/o  material knowledge across during tokenization training different number of tokens. } 
    \label{fig:MNMN} 
\end{figure}

\begin{table*}[ht]
\centering
\arraybackslash
\fontsize{8}{10.2}\selectfont 
\setlength{\tabcolsep}{3pt} 
\renewcommand{\arraystretch}{1.3} 
\resizebox{\textwidth}{!}{%
\begin{tabular}{c|cll|cll|cll}
\toprule
Tokenization & Concept & \multicolumn{1}{c}{Word Embedding} & \multicolumn{1}{c|}{Sim.} & Formula & \multicolumn{1}{c}{Word Embedding} & \multicolumn{1}{c|}{Sim.} & Abbr & \multicolumn{1}{c}{Word Embedding} & \multicolumn{1}{c}{Sim.} \\ \cmidrule{1-10}
\multirow{2}{*}{WordPiece} & \multirow{2}{*}{\begin{tabular}[c]{@{}c@{}}germanium\\ (german-ium)\end{tabular}} & agilent & 90.6 & \multirow{2}{*}{\begin{tabular}[c]{@{}c@{}}PbI2\\ (pib-2)\end{tabular}} & nowak & 81.8 & \multirow{2}{*}{\begin{tabular}[c]{@{}c@{}}LFP\\ (lf-p)\end{tabular}} & inlet & 95.5 \\
 &  & fri & 85.9 &  & 10c & 81.8 &  & chattopadhyay & 93.7 \\ \midrule 
\multirow{2}{*}{SAGE} & \multirow{2}{*}{\begin{tabular}[c]{@{}c@{}}germanium\\ (german-ium)\end{tabular}} & lot & 83.0 & \multirow{2}{*}{\begin{tabular}[c]{@{}c@{}}PbI2\\ (p-ib-2)\end{tabular}} & -gen & 43.9 & \multirow{2}{*}{\begin{tabular}[c]{@{}c@{}}LFP\\ (lf-p)\end{tabular}} & occupation & 95.8 \\
 &  & segregation & 82.8 &  & pounds & 43.6 &  & multiphonon & 95.2 \\ \midrule 
\multirow{2}{*}{PickyBPE} & \multirow{2}{*}{\begin{tabular}[c]{@{}c@{}}germanium\\ (g-erman-ium)\end{tabular}} & nomin & 81.2 & \multirow{2}{*}{\begin{tabular}[c]{@{}c@{}}PbI2\\ (p-bi-2)\end{tabular}} & gaussian & 63.1 & \multirow{2}{*}{\begin{tabular}[c]{@{}c@{}}LFP\\ (l-f-p)\end{tabular}} & her, & 75.8 \\
 &  & inex & 81.0 &  & p & 62.8 &  & consideration & 75.0 \\ \midrule 
\multirow{2}{*}{MATTER (ours)} & \multirow{2}{*}{\begin{tabular}[c]{@{}c@{}}germanium\\ (germanium)\end{tabular}} & \textbf{dithiocarbamate} & 81.5 & \multirow{2}{*}{\begin{tabular}[c]{@{}c@{}}PbI2\\ (pbi2)\end{tabular}} & \textbf{pb5} & 89.9 & \multirow{2}{*}{\begin{tabular}[c]{@{}c@{}}LFP\\ (lfp)\end{tabular}} & \textbf{zrf7} & 90.9 \\
 &  & \textbf{ammonium} & 81.4 &  & \textbf{pbf2} & 89.2 &  & \textbf{acyclohex} & 90.8 \\ \bottomrule
\end{tabular}
}
\caption{
Comparison of subword embedding averaging results across different tokenization methods.  The table presents the five nearest neighbor words based on subword embedding averages for each method. The similarity scores (Sim.) indicate the relevance of the nearest neighbors to the target material concept. \textbf{Boldface} highlights words that are directly related to materials.}
\label{tab:embedd}
\end{table*}

 \paragraph{Number of Tokens}
Figure \ref{fig:MNMN} presents the experimental results comparing MATTER with a tokenizer trained without material knowledge during tokenization training. The number of tokens was varied from 0.5x to 1.5x of the original size. The results show that MATTER consistently outperforms the tokenizer trained without material knowledge in all cases. This demonstrates that providing material-specific information during tokenization training is crucial, regardless of the token count.

\paragraph{Subword Embedding Analysis} Table~\ref{tab:embedd} presents the two nearest neighbors of material concepts using cosine similarity. The results show that the nearest neighbors of MATTER are more material-specific and semantically relevant compared to other methods. For instance, while WordPiece and SAGE generate less relevant neighbors (\textit{fri}, \textit{segregation} for \textit{germanium} \includegraphics[width=3.5mm]{figure/germanium.jpg}), our method produces material concepts such as \textit{dithiocarbamate} and \textit{ammonium} for \textit{germanium} \includegraphics[width=3.5mm]{figure/germanium.jpg}. This indicates our tokenizer better preserves material-specific meanings, improving representation quality for scientific text.

Further inspection reveals that the learned subword embeddings capture a variety of chemically meaningful relationships. For example, pairs such as \textit{PbI$_2$} and \textit{PbF$_2$} belong to the same chemical family of lead halides, while \textit{germanium} and \textit{dithiocarbamate} co-occur as known compound pairs in Ge–S coordination complexes. Other relationships reflect compositional connections, such as the coexistence of \textit{germanium} and \textit{ammonium} in ammonium tris(oxalato)germanate, or functional similarity, as seen in \textit{LFP} and \textit{ZrF$_7$}, both of which are used in energy storage and sensing applications.

These findings support the claim that the embedding space goes beyond capturing surface-level co-occurrence, instead reflecting deeper, domain-relevant semantics. A more comprehensive analysis and additional examples can be found in Appendix~\ref{sec:Subword_Embedding_Analysis}.

\begin{table*}[ht]
\centering
\arraybackslash
\fontsize{8}{10.2}\selectfont 
\setlength{\tabcolsep}{3pt} 
\renewcommand{\arraystretch}{1.3} 
\resizebox{\textwidth}{!}{%
\begin{tabular}{ccrrrrrrrr}
\toprule
 &  & \multicolumn{8}{c}{MatSci-NLP} \\ \cmidrule{3-10} 
\multirow{-2}{*}{Tokenization} & \multirow{-2}{*}{Metric} & \multicolumn{1}{c}{NER} & \multicolumn{1}{c}{RC} & \multicolumn{1}{c}{EAE} & \multicolumn{1}{c}{PC} & \multicolumn{1}{c}{SAR} & \multicolumn{1}{c}{SC} & \multicolumn{1}{c}{SF} & \multicolumn{1}{c}{Overall} \\ \midrule 
 & Micro-F1 & ${76.6}_\text{\textcolor{gray}{±0.2}}$ & ${80.9}_\text{\textcolor{gray}{±0.3}}$ & ${48.5}_\text{\textcolor{gray}{±0.2}}$ & ${73.1}_\text{\textcolor{gray}{±0.5}}$ & $\underline{81.9}_\text{\textcolor{gray}{±0.4}}$ & ${90.0}_\text{\textcolor{gray}{±0.1}}$ & ${57.4}_\text{\textcolor{gray}{±0.2}}$ & ${72.6}_\text{\textcolor{gray}{±0.1}}$ \\
\multirow{-2}{*}{\begin{tabular}[c]{@{}c@{}}w/o    material knowledge  \\ (WordPiece)\end{tabular}} & Macro-F1 & ${56.1}_\text{\textcolor{gray}{±0.2}}$ & ${58.5}_\text{\textcolor{gray}{±0.6}}$ & ${29.4}_\text{\textcolor{gray}{±0.3}}$ & ${58.9}_\text{\textcolor{gray}{±1.0}}$ & $\underline{74.6}_\text{\textcolor{gray}{±0.9}}$ & ${60.3}_\text{\textcolor{gray}{±0.8}}$ & ${32.6}_\text{\textcolor{gray}{±0.2}}$ & ${52.9}_\text{\textcolor{gray}{±0.2}}$ \\ \midrule  \midrule 
 & Micro-F1 & $\underline{77.1}_\text{\textcolor{gray}{±1.1}}$ & $\underline{81.5}_\text{\textcolor{gray}{±0.7}}$ & $\underline{53.1}_\text{\textcolor{gray}{±3.5}}$ & $\underline{73.6}_\text{\textcolor{gray}{±0.6}}$ & ${80.6}_\text{\textcolor{gray}{±3.3}}$ & $\underline{91.2}_\text{\textcolor{gray}{±1.0}}$ & $\underline{58.8}_\text{\textcolor{gray}{±2.5}}$ & $\underline{73.7}_\text{\textcolor{gray}{±1.3}}$ \\
\multirow{-2}{*}{\begin{tabular}[c]{@{}c@{}}ChemDataExtractor   \\ \cite{swain2016chemdataextractor}\end{tabular}} & Macro-F1 & $\underline{56.4}_\text{\textcolor{gray}{±1.3}}$ & $\underline{58.9}_\text{\textcolor{gray}{±0.7}}$ & $\underline{35.0}_\text{\textcolor{gray}{±4.2}}$ & $\underline{67.6}_\text{\textcolor{gray}{±1.2}}$ & ${68.0}_\text{\textcolor{gray}{±9.5}}$ & $\underline{64.8}_\text{\textcolor{gray}{±0.1}}$ & $\underline{35.6}_\text{\textcolor{gray}{±1.6}}$ & $\underline{55.2}_\text{\textcolor{gray}{±2.8}}$ \\ \midrule 
\rowcolor[HTML]{E9EBF5} 
\cellcolor[HTML]{E9EBF5} & Micro-F1 & $\textbf{80.0}_\text{\textcolor{gray}{±0.0}}$ & $\textbf{83.8}_\text{\textcolor{gray}{±0.1}}$ & $\textbf{53.1}_\text{\textcolor{gray}{±0.2}}$ & $\textbf{73.7}_\text{\textcolor{gray}{±0.2}}$ & $\textbf{85.5}_\text{\textcolor{gray}{±0.3}}$ & $\textbf{91.2}_\text{\textcolor{gray}{±0.1}}$ & $\textbf{61.9}_\text{\textcolor{gray}{±0.3}}$ & $\textbf{75.6}_\text{\textcolor{gray}{±0.1}}$ \\
\rowcolor[HTML]{E9EBF5} 
\multirow{-2}{*}{\cellcolor[HTML]{E9EBF5}MatDetector (ours)} & Macro-F1 & $\textbf{59.3}_\text{\textcolor{gray}{±0.2}}$ & $\textbf{59.1}_\text{\textcolor{gray}{±0.5}}$ & $\textbf{36.9}_\text{\textcolor{gray}{±0.3}}$ & $\textbf{67.6}_\text{\textcolor{gray}{±0.6}}$ & $\textbf{79.3}_\text{\textcolor{gray}{±0.7}}$ & $\textbf{64.9}_\text{\textcolor{gray}{±0.5}}$ & $\textbf{38.0}_\text{\textcolor{gray}{±0.3}}$ & $\textbf{57.9}_\text{\textcolor{gray}{±0.1}}$ \\ \bottomrule

\end{tabular}
}

  \caption{Ablation results on different detectors for the MatSci-NLP dataset across multiple tasks. w/o  material knowledge represents frequency-centric tokenization without any additional signal. ChemDataExtractor and MatDetector incorporate additional signals using their respective tools. \textbf{Bold} values indicate the highest scores for each metric-task pair, while \underline{underline} represent the second-highest scores.}
  \label{tab:ab}
\end{table*}
\subsection{Ablation Study} 
\label{sec:ab}

\paragraph{Comparison of Detectors} To confirm whether using MatDetector to extract material concepts and assign weights is more suitable for providing accurate and domain-relevant signals in the material domain compared to the widely used ChemDataExtractor, we performed ablation studies. Specifically, we replaced MatDetector with ChemDataExtractor to assign weights. While ChemDataExtractor is capable of partially extracting material concepts, it lacks the ability to assess the importance of the extracted concepts within the material domain. Consequently, all material concepts extracted by ChemDataExtractor were assigned the highest signal weight of 0.99.

Table \ref{tab:ab} show that using MatDetector outperforms ChemDataExtractor, achieving a 2\% higher average Micro-F1 score and a 2.7\% higher Macro-F1 score. This confirms that MatDetector is more effective in providing material domain-relevant signals. Additionally, when examining the performance of ChemDataExtractor, we observed that it achieved a 1.1\% higher Micro-F1 score and a 2.3\% higher Macro-F1 score compared to the baseline method, which did not incorporate any material signals. This underscores the importance of incorporating material signals into tokenization.

\begin{figure}[t] 
    \centering 
    \includegraphics[width=0.48\textwidth]{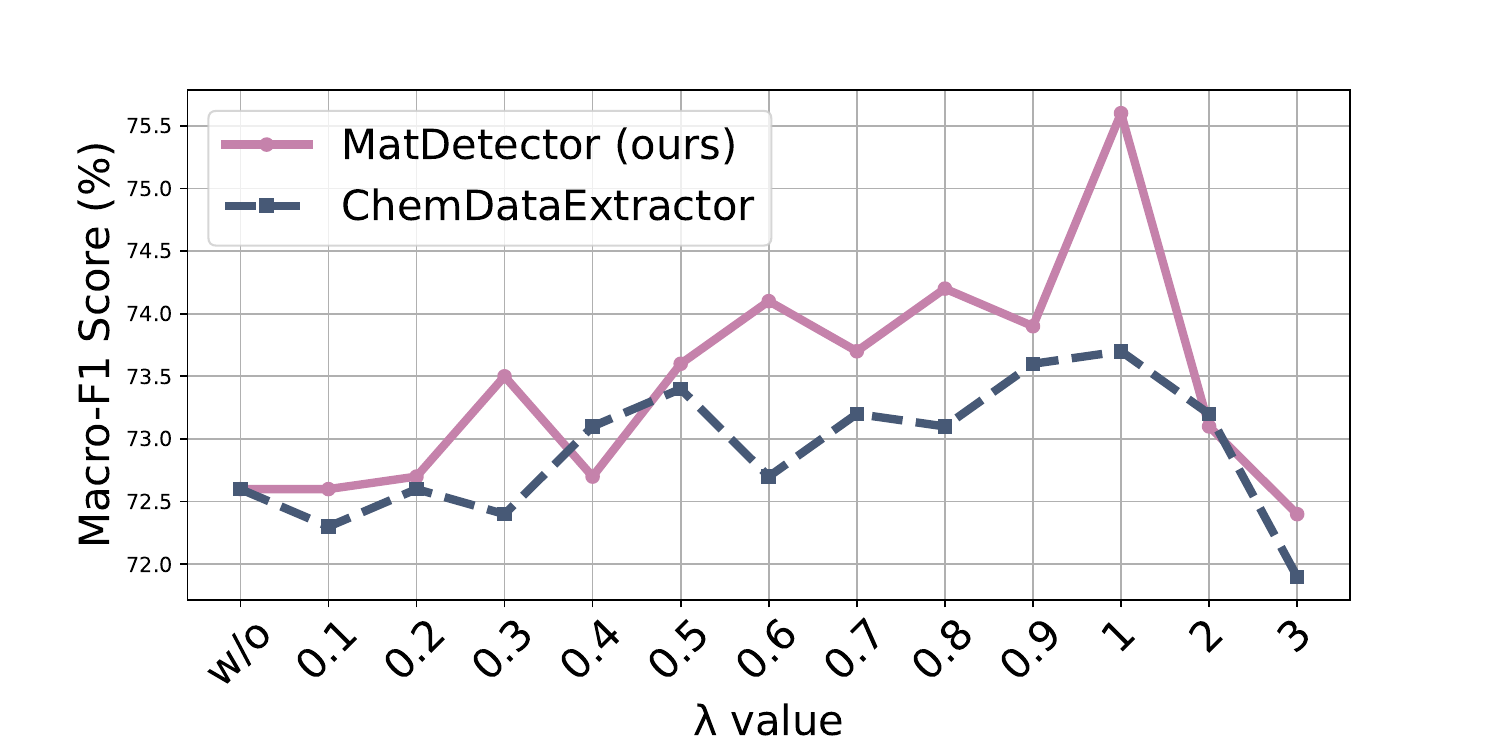} 
    \caption{Comparison of Macro-F1 scores for ChemDataExtractor and MatDetector across $\lambda$ values. } 
    \label{fig:MDF1} 
\end{figure}

However, as evidenced by the performance gap between ChemDataExtractor and MatDetector, it is clear that the accuracy of the material signals plays a critical role. These results highlight the necessity of not only incorporating material signals but also ensuring that accurate material concepts and their respective significance are properly considered. The use of MatDetector effectively addresses both aspects, demonstrating its suitability for enhancing performance in the material domain. Both detectors achieved their highest performance at a $\lambda$ value of $1$ as show in Figure \ref{fig:MDF1}.

\paragraph{Comparison of Lambda} Figure \ref{fig:MDF1} demonstrates that adding material signals, regardless of the weighting method used, consistently yields better performance compared to the baseline where no material signals were incorporated (\( \lambda \) =0). This observation aligns with previous findings and further substantiates that the inclusion of material Knowledge is beneficial. Moreover, it emphasizes the necessity of using appropriate tools to effectively assign these signals for optimal performance.

Notably, both ChemDataExtractor and MatDetector achieved their highest performance at $\lambda = 1$. Based on this consistent observation across models, all preceding experiments in this study were conducted using this optimal setting.

\section{Conclusion}
We proposed MATTER, a novel tokenization approach that incorporates material knowledge derived from material corpora into the tokenization process. MATTER has enabled the creation of vocabularies tailored to the material domain, effectively maintaining the structure and semantics of material concepts. Our extensive experiments have demonstrated that MATTER tokenization significantly improves performance across a wide range of material generation and classification tasks, outperforming conventional tokenization methods. Our work has provided a strong, adaptable foundation components for materials NLP, empowering future research on materials science.

\section*{Limitations}
While we have demonstrated that MATTER effectively enhances tokenization for pretrained language models in the materials science domain. Nevertheless, our work also opens several valuable opportunities for further improvements and exploration.

\paragraph{Hyperparameter Selection.}  MATTER introduces a tunable hyperparameter (\( \lambda \)) to balance frequency statistics with material-specific signals during vocabulary construction. While we observed stable improvements across a range of \( \lambda \) values, the method still requires manual selection of this parameter. Although \( \lambda = 1 \) was found to be effective in our experiments, identifying an optimal value for different domains or corpora may require additional tuning. This reliance on hyperparameter selection may affect general usability in practice.

\paragraph{Further Analysis on Corpus}  
The current experiments were conducted following the prior methodology outlined in \cite{gupta2022matscibert}, which emphasizes the use of material-specialized corpora. Although this ensures consistency and relevance to domain-specific evaluation, future work may benefit from expanding the diversity of training corpora to test MATTER’s generalizability across subdomains and heterogeneous sources.

\paragraph{NER Dependency and Scalability}  
Our approach relies on the identification of material concepts through NER-based classification. To support this, we constructed a high-quality NER dataset using a curated materials knowledge base, ensuring accurate detection of domain-specific terminology essential for effective vocabulary construction in materials science. However, this reliance on supervised signals may introduce challenges in scalability, particularly when applied to broader or less-structured corpora. Addressing this limitation remains an important direction for future work.

\section*{Acknowledgements}
This work was supported by the National Research Foundation of Korea (NRF) grant funded by the Korea government (MSIT) (No.RS-2025-00517221 and No.RS-2024-00415812) and Institute of Information \& communications Technology Planning \& Evaluation (IITP) grant funded by the Korea government (MSIT) (No.RS-2024-00439328, Karma: Towards Knowledge Augmentation for Complex Reasoning (SW Starlab), No.RS-2024-00457882, AI Research Hub Project, and No.RS-2019-II190079, Artificial Intelligence Graduate School Program (Korea University)).

\bibliography{custom}
\clearpage

\appendix

\section{Implementation Details and Setups}
\label{sec:Implementation}

\subsection{Tokenization baseline}
\label{sec:Implementationb1}

We compared our tokenization approach against baseline methods, including the widely used frequency-centric tokenization, WordPiece, as well as more recent and strong tokenization methods, SAGE and PickyBPE. To ensure a fair comparison, all tokenization methods adhered to the vocabulary size of 31,090, as defined in the prior methodology \cite{gupta2022matscibert}. The implementation details are as follows:

\paragraph{WordPiece \cite{wu2016google}}  being one of the most widely used and fundamental frequency-centric tokenization methods, was configured with a min frequency of 2 and a limit alphabet of 6,000. 

\paragraph{SAGE \cite{yehezkel2022incorporating} } enhances frequency-centric tokenization by incorporating contextual signals into the process. The implementation of SAGE included several key parameters:  vocabulary schedule progressively reducing from 32,000 to the target size of 31,090; an embedding schedule synchronized with the vocabulary schedule; a maximum token length of 17 bytes; and the use of skip-gram embedding training with a vector size of 50, a context window size of 5, and 15 negative samples. To ensure reproducibility, the random seed was set to 692,653.

\paragraph{PickyBPE  \cite{chizhov2024bpe}} was employed to construct the vocabulary, with a desired vocabulary size of 31,090 and an IoS (Importance of Symbols) threshold set to 0.9. The initial vocabulary ensured comprehensive coverage with a relative symbol coverage of 0.9999. During training, the frequency of merges was logged at intervals of 200 merges to monitor the tokenization process effectively.

\subsection{Hyper-parameters}
\label{sec:Implementationb2}

\paragraph{Pre-Train.} 

We follow the previous work \cite{gupta2022matscibert}. The detailed configuration of the main model and training hyperparameters is summarized as follows:
\begin{table}[H]
\centering
\resizebox{0.45\textwidth}{!}{ 
\begin{tabular}{cc}
\toprule
Parameter & Value \\ \midrule 
Encoder   Layers & 12 \\
Embedding   Dim & 768 \\
Hidden   Dim & 768 \\
Attention   Heads & 12 \\ \midrule 
Max   Tokens in a Batch & 128 \\
Optimizer & Adam \\
Weight   Decay & 0.01 \\
Learning   Rate (LR) & 2e-5 \\
LR   Scheduler & Linear with Warmup \\
Warmup   Strategy & Linear \\
Precision & FP16 \\ \bottomrule
\end{tabular}
}

  \label{tab:pre_hp}
\end{table}

\begin{table*}[ht]
\centering
\begin{tabular}{cccccc}
\toprule
\multirow{2}{*}{Parameter} & \multicolumn{5}{c}{{Value}} \\ \cmidrule{2-6} 
 & NER$_\text{SOFC}$ & NER$_\text{Matscholar}$ & SF & RC & PC \\ \midrule 
Max   Tokens in a Batch & 32 & 32 & 32 & 64 & 64 \\
Training   Epochs & 20 & 15 & 40 & 10 & 10 \\ 
Optimizer & \multicolumn{5}{c}{Adam} \\
Learning   Rate (LR) & \multicolumn{5}{c}{[2e-5, 3e-5, 5e-5]} \\
LR   Scheduler & \multicolumn{5}{c}{Linear with Warmup} \\
Warmup   Strategy & \multicolumn{5}{c}{Warmup ratio of 0.1} \\
Precision & \multicolumn{5}{c}{FP32} \\
\bottomrule
\end{tabular}
\caption{Detailed configuration of the main model and training hyperparameters for classification task.} 
  \label{tab:cls_hp}
\end{table*}
\subsection{Evaluation metrics} 
\label{sec:Implementationbem}

\paragraph{Classification task.} 
We follow the previous work \cite{gupta2022matscibert}. The detailed configuration of the main model and training hyperparameters is summarized in Table \ref{tab:cls_hp}.

\paragraph{Generation task.} 

We follow the previous work \cite{song2023matsci}. The detailed configuration of the main model and training hyperparameters is summarized as follows:

\begin{table}[H]
\centering
\resizebox{0.45\textwidth}{!}{ 
\begin{tabular}{cc}
\toprule
Parameter & Value \\ \midrule 
Decoder   Layers & 3 \\
Embedding   Dim & 768 \\
Hidden   Dim & 768 \\
Attention   Heads & 8 \\ \midrule 
Max   Tokens in a Batch & 4 \\
Optimizer & Adam \\
Learning   Rate (LR) & 2e-5 \\
Precision & FP32 \\
Training   Epochs & \begin{tabular}[c]{@{}c@{}}Up to 20 \\ (early stopping)\end{tabular} \\ \bottomrule
\end{tabular}
}
\end{table}

\begin{table*}[ht]
\centering
\resizebox{0.9\textwidth}{!}{ 
\begin{tabular}{
>{\columncolor[HTML]{FFFFFF}}l 
>{\columncolor[HTML]{FFFFFF}}l |
>{\columncolor[HTML]{FFFFFF}}c 
>{\columncolor[HTML]{FFFFFF}}c }
\toprule
\multicolumn{2}{c|}{\cellcolor[HTML]{FFFFFF}{\color[HTML]{000000} {Tool}}} & {\color[HTML]{000000} {\begin{tabular}[c]{@{}c@{}}ChemDataExtractor\\ \cite{swain2016chemdataextractor}\end{tabular}}} & {\color[HTML]{000000} {MatDetector (ours)}} \\ \midrule 
\multicolumn{2}{l|}{\cellcolor[HTML]{FFFFFF}{\color[HTML]{000000} Entity type}} & {\color[HTML]{000000} chemical mention} & {\color[HTML]{000000} chemical concepts, formulas} \\ \midrule 
\cellcolor[HTML]{FFFFFF}{\color[HTML]{000000} } & {\color[HTML]{000000} domain} & {\color[HTML]{000000} BioMedical} & {\color[HTML]{000000} Material} \\
\cellcolor[HTML]{FFFFFF}{\color[HTML]{000000} } & {\color[HTML]{000000} annotated method} & {\color[HTML]{000000} manually annotated} & {\color[HTML]{000000} manually \& automatic annotated} \\
\multirow{-3}{*}{\cellcolor[HTML]{FFFFFF}{\color[HTML]{000000} Train data}} & {\color[HTML]{000000} number of abstract} & {\color[HTML]{000000} 10,000} & {\color[HTML]{000000} 404,262} \\ \bottomrule
\end{tabular}
}
\caption{Comparison of Extractable Entity Types and Training Data in ChemDataExtractor and MatDetector.}
\label{tab:ChemMat}
\end{table*}

We evaluate using metrics from MatSciNLP \cite{song2023matsci} and MatSciBERT \cite{gupta2022matscibert}. Generation tasks use Micro-F1 and Macro-F1, averaged over five seeds. Classification tasks report Macro-F1 (SOFC-NER, SOFC-Filling), Micro-F1 (MatScholar), and accuracy (Glass Science), with cross-validation over five folds and three seeds.

\subsection{Token Qualities Details}
\label{sec:qualieites_setting}
Among 31,090 vocabulary entries, we extract material-related tokens using MatDetector and compare tokenization methods.

\begin{table}[H]
\centering

\resizebox{0.35\textwidth}{!}{ 
\begin{tabular}{lc}
\toprule
Tokenization & \#material token \\ \midrule 
WordPiece & 10,420 \\
SAGE & 9,602 \\
PickyBPE & 9,203 \\
MATTER (ours) & \textbf{10,633} \\ \bottomrule
\end{tabular}
}
\end{table}

\section{Details of MatDetector Construction}
\label{sec:MatDetector}
MatDetector is a domain-specific Named Entity Recognition (NER) tool designed to extract material concepts from scientific texts. The detailed steps for its construction are as follows:

\paragraph{Crawling Material Corpus}
To construct the training dataset for the MatDetector, we first extract chemical names, IUPAC names, synonyms, and molecular formulas from PubChem \cite{kim2019pubchem}, obtaining 80K material concepts.  The number of concepts by category is provided in Table \cite{kim2019pubchem}. Using these extracted concepts as keywords, we collect 42K scientific papers from Semantic Scholar \cite{ammar2018construction}, focusing on titles and abstracts that contain high-density  material knowledge, with detailed comparative information in Table \ref{tab:ChemMat}. 

\paragraph{Creating Train Dataset}
While Semantic Scholar provides relatively clean data, most material-related data is collected from various journals and repositories, where formatting inconsistencies, OCR errors, and structural variations introduce significant noise. To address this, we construct a Noisy NER Dataset, improving model robustness and expanding the dataset to be four times larger than the original. The details of noise augmentation are as follow:
    \begin{itemize}
        \item \textbf{Material Name Noise}: This includes capitalization errors in element symbols, misplaced or duplicated digits, reordering of elements, and insertion of unnecessary characters or special symbols. These modifications reflect common errors found in chemical names and mimic the inconsistencies in scientific documents.
        \item \textbf{Material Formula Noise}: Common formatting inconsistencies in formulas are simulated by adding spaces around special symbols such as \verb|(|, \verb|)|, \verb|[|, and \verb|]|, or by replacing digits with placeholders. Combined patterns are also introduced to replicate multiple error types.
    \end{itemize}
Using this dataset, we generate a material NER dataset by tagging the collected corpus with material concepts extracted from PubChem, ensuring precise identification of material-related terminology. In this tagging process, \textit{Material Name}, IUPAC Name, and Synonym of Material Name are categorized as Material Concept, while Material Formula is tagged separately as \textit{Material Formula}. This approach maintains a clear distinction between conceptual material entities and their chemical formulas, enabling more accurate entity recognition in materials science applications.

\paragraph{Training the MatDetector}
We train the MatDetector using the material NER dataset constructed in the previous step and the \citet{trewartha2022quantifying} model architecture. The model achieves high accuracy in detecting material concepts, even in noisy corpora, and provides NER tagging probabilities, estimating the likelihood that a concepts belongs to materials science.

\section{Additional QA Experiments on MaScQA}
\label{sec:qa-exp}

To evaluate the generalizability of MATTER beyond classification tasks, we conducted additional experiments on the MaScQA \cite{zaki2024mascqa} benchmark, which focuses on materials-domain question answering.

\paragraph{Decoder-based setup.}  
We fine-tuned two decoder-based models— Llama-3.2-1B-Instruct and SciBERT—on the HoneyBEE~\cite{song2023honeybee} instruction dataset and evaluated their performance on MaScQA. MATTER consistently achieved higher accuracy compared to other tokenizations:

\begin{table}[h]
\centering
\begin{tabular}{l|c}
\toprule
Tokenization & {Accuracy (\%)} \\
\midrule 
BPE & 7.1 \\
PickyBPE & 7.3 \\
MATTER (ours) & \textbf{8.9} \\
\bottomrule
\end{tabular}
\caption{MaScQA benchmark accuracy using decoder-based models.}
\end{table}

\paragraph{Encoder-decoder setup.}  
Following the setup in MatSciNLP~\cite{song2023matsci}, we used MatSciBERT as the encoder and a transformer-based decoder. We trained on 10\% of the HoneyBEE QA data and evaluated on the remaining 90\%, simulating a low-resource QA scenario. MATTER again yielded the best performance:

\begin{table}[h]
\centering

\begin{tabular}{l|c}
\toprule
Tokenization & {Accuracy (\%)} \\
\midrule 
BPE & 20.01 \\
WordPiece & 22.74 \\
SAGE & 22.93 \\
PickyBPE & 21.01 \\
MATTER (ours) & \textbf{23.96} \\
\bottomrule
\end{tabular}
\caption{MaScQA benchmark performance with encoder-decoder model.}
\end{table}

These results confirm MATTER’s effectiveness in enhancing QA performance across diverse model architectures and reinforce its generalizability to downstream materials tasks.

\section{Statistical Significance}
\label{sec:Statistical_Significance}

\paragraph{Generation task} To quantitatively assess the statistical significance of performance improvements introduced by {MATTER}, we conducted paired t-tests on the \textit{average F1 scores} across eight generation tasks (NER, RC, EAE, PC, SAR, SC, SF, Overall), comparing MATTER against four widely-used tokenization baselines: {BPE}, {WordPiece}, {SAGE}, and {PickyBPE}. The average F1 score was computed as the arithmetic mean of the Micro-F1 and Macro-F1 values for each task.

The paired t-test evaluates whether the mean difference in Avg-F1 scores between MATTER and a baseline is statistically significant. The t-statistic is given by:

\begin{equation}
t = \frac{\bar{d}}{s_d / \sqrt{n}}
\end{equation}

where $\bar{d}$ is the mean of the differences between MATTER and a baseline across tasks, $s_d$ is the standard deviation of those differences, and $n = 8$ is the number of generation tasks.

\vspace{1em}
\begin{table}[h]
\centering
\renewcommand{\arraystretch}{1.2}
\begin{tabular}{lcc}
\toprule
{Tokenization} & {Avg-F1 ($p$)} & {Significant} \\
\midrule 
BPE         & 0.0009 & Yes \\
WordPiece   & 0.0001 & Yes \\
SAGE        & 0.0066 & Yes \\
PickyBPE    & 0.0155 & Yes \\

\bottomrule
\end{tabular}
\caption{Paired t-test results comparing the average F1 score between MATTER and each baseline across generation tasks.}

\label{tab:ttest_gen_avgf1}
\end{table}

As shown in Table~\ref{tab:ttest_gen_avgf1}, {MATTER achieves statistically significant improvements} over all four baselines in terms of average F1 score. All comparisons yield $p < 0.05$, confirming that MATTER’s performance gains are unlikely due to random variation. These results reinforce the effectiveness of MATTER’s domain-aware tokenization strategy in improving generation performance across diverse material-related tasks.

\paragraph{Classification task} We conducted the same analysis for classification tasks to evaluate whether MATTER’s improvements generalize to discriminative settings. Paired t-tests were performed on the \textit{average F1 scores} across five classification tasks (SOFC-NER, MatScholar-NER, SF, RC, PC), using the same computation.

\vspace{1em}
\begin{table}[h]
\centering
\renewcommand{\arraystretch}{1.2}
\begin{tabular}{lcc}
\toprule
{Tokenization} & {Avg-F1 ($p$)} & {Significant} \\
\midrule 
BPE         & 0.0001 & Yes \\
WordPiece   & 0.0001 & Yes \\
SAGE        & 0.0021 & Yes \\
PickyBPE    & 0.0009 & Yes \\
\bottomrule
\end{tabular}
\caption{Paired t-test results comparing the average F1 score between MATTER and each baseline across classification tasks.}
\label{tab:ttest_cls_avgf1}
\end{table}

As shown in Table~\ref{tab:ttest_cls_avgf1}, all comparisons again yield statistically significant results ($p < 0.005$), confirming that {MATTER consistently outperforms all baselines} in overall classification performance. This aggregated F1-based analysis further demonstrates the robustness of MATTER’s tokenization advantages in both generation and classification tasks, effectively balancing frequency-weighted and class-balanced evaluation perspectives.

\section{Details of validation on materials NER}
\label{sec:NER}

\renewcommand{\arraystretch}{1.5}  
\begin{table}[H]
\centering
\setlength{\tabcolsep}{6pt}  
{\footnotesize
\begin{tabular}{l|ccc}
\toprule
\multicolumn{1}{l|}{Tool} & \multicolumn{1}{l}{MatScholar} & \multicolumn{1}{l}{SOFC} & \multicolumn{1}{l}{Overall} \\ \midrule 
ChemDataExtractor & 12\% & 24\% & 18\% \\
MatDetector (ours) & \textbf{53\%} & \textbf{60\%} & \textbf{57\%} \\ \bottomrule
\end{tabular}
}
\caption{Recall of two material concept extraction tools on external materials NER datasets—MatScholar and SOFC.}
\label{tab:nerner_re1}
\end{table}

\begin{table}[H]
\centering
\setlength{\tabcolsep}{6pt}  
{\footnotesize
\begin{tabular}{l|ccc}
\toprule
\multicolumn{1}{l|}{Tool} & \multicolumn{1}{l}{MatScholar} & \multicolumn{1}{l}{SOFC} & \multicolumn{1}{l}{Overall} \\ \midrule 
ChemDataExtractor & 52\% & 61\% & 57\% \\
MatDetector (ours) & \textbf{63\%} & \textbf{75\%} & \textbf{69\%} \\ \bottomrule
\end{tabular}
}
\caption{Precision of two material concept extraction tools on external materials NER datasets—MatScholar and SOFC.}
\label{tab:nerner_re2}
\end{table}

\begin{table}[H]
\centering
\setlength{\tabcolsep}{6pt}  
{\footnotesize
\begin{tabular}{l|ccc}
\toprule
\multicolumn{1}{l|}{Tool} & \multicolumn{1}{l}{MatScholar} & \multicolumn{1}{l}{SOFC} & \multicolumn{1}{l}{Overall} \\ \midrule 
ChemDataExtractor & 20\% & 34\% & 27\% \\
MatDetector (ours) & \textbf{58\%} & \textbf{67\%} & \textbf{63\%} \\ \bottomrule
\end{tabular}
}
\caption{F1 Score of two material concept extraction tools on external materials NER datasets—MatScholar and SOFC.}
\label{tab:nerner_re3}
\end{table}

\section{Details of the Word-Initial Token Analysis}
\label{sec:Details_Word-Initial_Token}
To validate the effectiveness of our tokenization and avoid any potential circularity in evaluation, we perform an additional analysis using external and independent sources of material-related terms, separate from those used to construct the tokenization. Specifically, we collect named entities from two manually annotated materials NER datasets used in the paper:

\begin{table}[H]
\centering
\setlength{\tabcolsep}{6pt}  
{\footnotesize
\begin{tabular}{lc}
\toprule
NER Dataset & \#Material Entity \\ \midrule 
MatScholar \cite{weston2019named} & 8,660 \\
SOFC \cite{friedrich2020sofc} & 1,201 \\ \midrule 
Total & 9,861 \\ \bottomrule
\end{tabular}
}

\label{tab:nerner_re}
\end{table}

\section{Case Study: Tokenization Robustness}
\label{sec:Tokenization_Robustness}

\subsection{Analysis in Material Science Papers} 
In this section, we applied WordPiece, SAGE, PickyBPE, and the proposed method, MATTER, to tokenization results from real materials science papers.
As shown in Table \ref{tab:tokenization_results_real}, existing tokenization methods such as WordPiece, SAGE, and PickyBPE tend to overtokenize important material concepts. For instance, the chemical formula for Lead, "Pb", is split into "p-b", while "dimethylsiloxane" is divided into "dimethyl-sil-oxane or d-imethyl-sil-oxane". Such overtokenization distorts the semantic integrity of material concepts and can degrade the performance of downstream natural language processing tasks.

In contrast, our proposed MATTER method effectively prevents the overtokenization of material concepts. When applying MATTER, essential material concepts such as "Pb", "dimethylsiloxan"e, and "barium sulfate" remain intact, preserving their contextual meaning. Notably, complex material concepts such as "perovskite" and "ethylene-diaminetetraacetic acid" are properly maintained, demonstrating that MATTER provides a more suitable tokenization approach for materials science texts.

\subsection{Subword Embedding Analysis} 
\label{sec:Subword_Embedding_Analysis}
To evaluate the impact of different tokenization methods on word representations in materials science, we analyze the nearest neighbors of material concepts based on subword embedding averaging. This experiment is conducted in conjunction with the tokenization results presented in Figure ~\ref{fig:1_graph} and Table \ref{tab:tokenization_results_real}, allowing us to assess how subword segmentation affects semantic consistency in word embeddings. We compare four tokenization strategies—WordPiece, SAGE, PickyBPE, and our proposed method, MATTER—by computing word embeddings as the mean of their constituent subword embeddings. The similarity between words is measured using cosine similarity, and the five nearest neighbors (5-NN) for each concept are retrieved. The retrieved neighbors allow us to assess whether the tokenization method preserves materials science semantics or introduces artifacts from suboptimal subword segmentation. The dataset used for evaluation includes materials science terminology, chemical formulas, and domain-specific abbreviations, ensuring a realistic assessment of tokenization impact.

The results, presented in Table \ref{tab:sim top5}, indicate that WordPiece and SAGE exhibit a strong tendency to retrieve words that share surface-level subword structures rather than those with true material relevance. For instance, 'germanium' is tokenized as german-ium in WordPiece, leading to nearest neighbors such as 'german' and '-ium', which lack meaningful chemical associations. PickyBPE partially alleviates this issue by merging frequent subwords, but still retrieves words that reflect tokenization artifacts rather than conceptually related material concepts. In contrast, our MATTER method significantly improves semantic alignment by retrieving chemically relevant words. For example, the nearest neighbors of 'germanium' include 'dithiocarbamate', 'ammonium', and 'borohydride', demonstrating a stronger connection to materials science concepts. Similarly, 'ethylenediaminetetra-acetic' acid retrieves '-oxycarb' and -'sulfanyl', which accurately reflect its chemical properties. These results suggest that MATTER effectively mitigates tokenization-induced distortions, leading to more precise materials science word representations that enhance performance in downstream NLP tasks such as entity linking, material property prediction, and knowledge graph construction.

\begin{table}[t] 
\centering
\renewcommand{\arraystretch}{1.1} 
\begin{tabular}{@{}lc@{}}
\toprule
Type of material concept & \#material concept     \\ \midrule
Material name            & 22,482  \\
IUPAC name               & 22,482  \\
Synonym of material name & 719,885 \\
Material formula         & 22,479  \\ \bottomrule
\end{tabular}
\caption{Summary of approximately 80K extracted material concepts from PubMed, categorized by concepts type.}
\label{tab:material_terms}
\end{table}
\subsection{Comparison of Lambda Details}
\label{sec:Lambda}
\begin{figure}[t] 
    \centering 
    \includegraphics[width=0.48\textwidth]{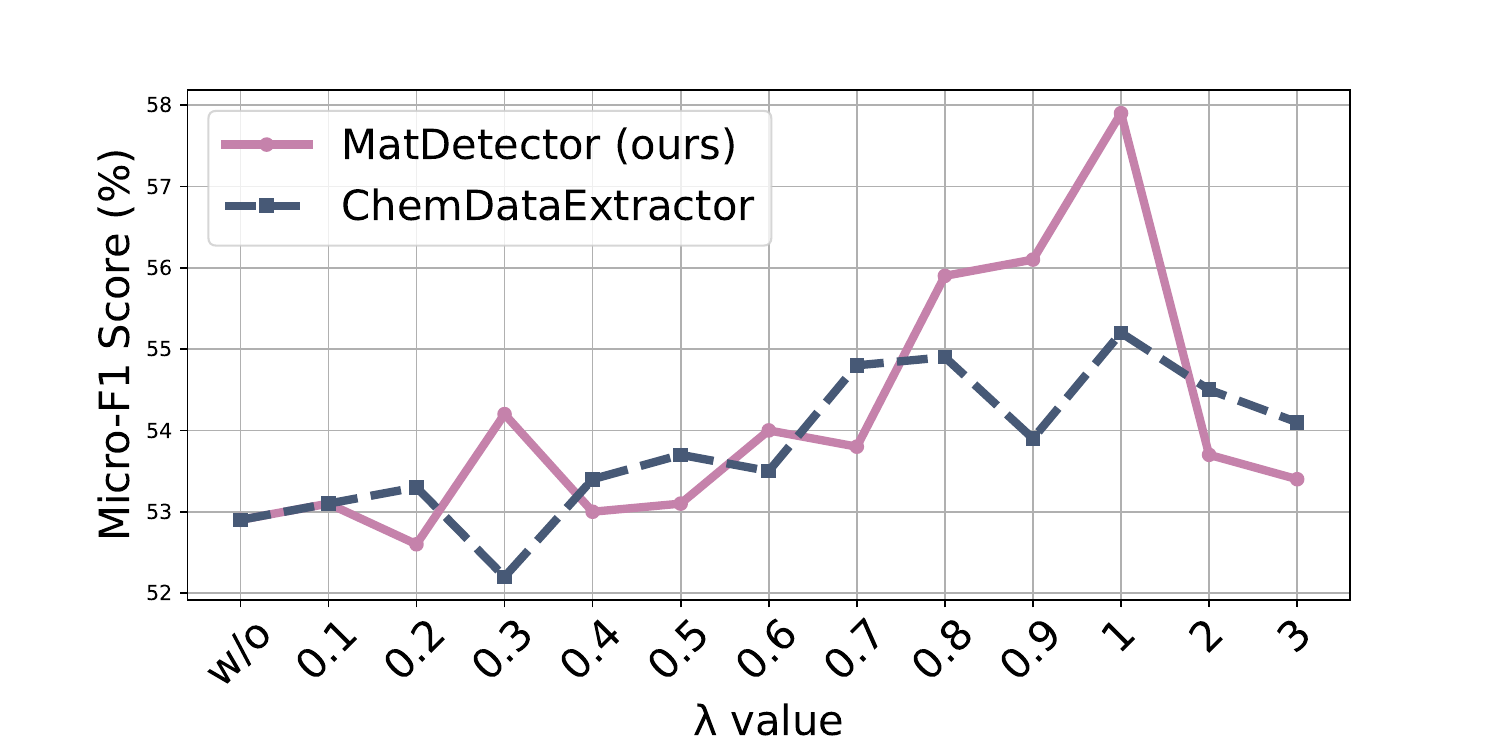} 
    \caption{Comparison of Micro-F1 scores for ChemDataExtractor and MatDetector across different $\lambda$ values. } 
    \label{fig:MD} 
\end{figure}

The Macro-F1 scores for ChemDataExtractor and MatDetector were compared across different $\lambda $ values to evaluate their performance. The specific numerical values are detailed in Table \ref{tab:ab1} and Table \ref{tab:ab2}, while Figure \ref{fig:MD} provides a visual representation for easier interpretation.

\begin{table*}[ht]
    \centering
    \renewcommand{\arraystretch}{1.2} 
    
    \begin{tabular}{l p{12cm}} 
        \hline
        \textbf{Method} & \textbf{Tokenized Output} \\ \hline
    Origin & \textcolor[HTML]{db5079}{\textbf{poly}} ( \textcolor[HTML]{db5079}{\textbf{dimethylsiloxane}} ) ( \textcolor[HTML]{db5079}{\textbf{pdms}} ) waschosen to form the \textcolor[HTML]{db5079}{\textbf{elastomeric}} circuit board, mechanically protecting and electrically insulating the wires, based on itsdurability, adjustable stiffness, \textbf{biocompatibility}, and commer-cial availability as an insulating compound. \cite{gray2004high} \\ \hline
        WordPiece & \textcolor[HTML]{db5079}{\textbf{poly}} ( \textcolor[HTML]{db5079}{\textbf{dimethyl-sil-oxane}} ) ( \textcolor[HTML]{db5079}{\textbf{pd-ms}} ) was-cho-sen to form the \textcolor[HTML]{db5079}{\textbf{elast-omeric}} circuit board , mechanically protecting and electrically ins-ulating the wires , based on its-du-rab-ility , adjustable stiffness , \textbf{bioc-omp-ati-bility} , and comme-r-ci-al availability as an ins-ulating compound. \\
        SAGE & \textcolor[HTML]{db5079}{\textbf{poly}} ( \textcolor[HTML]{db5079}{\textbf{dimethyl-siloxane}} ) ( \textcolor[HTML]{db5079}{\textbf{pd-ms}} ) was-cho-sen to form the \textcolor[HTML]{db5079}{\textbf{elastomer-ic}} circuit board , mechanically protecting and electrically insulating the wires , based on its-du-rab-ility , adjustable stiffness , \textbf{biocompatibility} , and comme-r-ci-al availability as an insulating compound. \\
        PickyBPE & \textcolor[HTML]{db5079}{\textbf{p-oly}} ( \textcolor[HTML]{db5079}{\textbf{d-imethyl-sil-xane}} ) ( \textcolor[HTML]{db5079}{\textbf{p-d-ms}} ) was-ch-osen to form the \textcolor[HTML]{db5079}{\textbf{elast-omeric}} circuit board , mechanically protecting and electrically insulating the wires , based on its-d-urability , adjustable stiffness , \textbf{biocompatibility} , and comm-er-c-ial availability as an insulating compound. \\
        MATTER (ours) & \textcolor[HTML]{db5079}{\textbf{poly}} ( \textcolor[HTML]{db5079}{\textbf{dimethyl-siloxane}} ) ( \textcolor[HTML]{db5079}{\textbf{pdms}} ) was-cho-sen to form the \textcolor[HTML]{db5079}{\textbf{elastomeric}} circuit board , mechanically protecting and electrically insulating the wires , based on its-du-rab-ility , adjustable stiffness , \textbf{biocompatibility} , and comme-r-ci-al availability as an insulating compound. \\ \hline
        \hline
        Origin & t
        he waste was solubilized using \textcolor[HTML]{db5079}{\textbf{ethylenediaminetetraacetic acid}}, and its constituents were determined employing x-ray diffraction and inductively coupled \textbf{plasma-atomic} emission \textbf{spectrometry}, identifying \textbf{barium} sulfate ( \textcolor[HTML]{db5079}{\textbf{baso4}} ) as the predominant component at a weight percentage of 67.13\%. \cite{zhang2025efficiency}\\ \hline
        WordPiece & the waste was solubil-ized using \textcolor[HTML]{db5079}{\textbf{ethylenedi-amine-tetr-aa-ce-tic acid}} , and its constituents wer determined employing x - ray diffraction and inductively coupled \textbf{plasma - atomic} emission \textbf{spectrometry} , identifying \textbf{bari-um} sulfate ( \textcolor[HTML]{db5079}{\textbf{bas-o-4}} ) as the predominant component at a weight percentage of 67 . 13 \%. \\
        SAGE & the waste was solub-ilized using \textcolor[HTML]{db5079}{\textbf{ethylene-diam-inet-etra-ace-tic acid}} , and its constituents were determined employing x - ray diffraction and inductively coupled \textbf{plasma - atomic} emission \textbf{spectrometry} , identifying \textbf{barium} sulfate ( \textcolor[HTML]{db5079}{\textbf{bas-o-4}} ) as the predominant component at a weight percent-ag-e of 67 . 13 \%.\\
        PickyBPE & The waste was solub-ilized using \textcolor[HTML]{db5079}{\textbf{ethyl-ened-i-amin-et-etra-acetic acid}} , and its constituents were determined employing x - ray diffraction and inductively coupled \textbf{plasma - atomic} emission \textbf{spectrometry} , identifying \textbf{barium} sulfate ( \textcolor[HTML]{db5079}{\textbf{b-as-o-4}} ) as the predominant component at a weight percentage of 67.13 \%. \\
        MATTER (ours) & the waste was solub-ilized using \textcolor[HTML]{db5079}{\textbf{ethylenediaminetetra-acetic acid}} , and its constituents were determined employing x - ray diffraction and inductively coupled \textbf{plasma - atomic} emission \textbf{spectrometry} , identifying \textbf{barium} sulfate ( \textcolor[HTML]{db5079}{\textbf{baso4}} ) as the predominant component at a weight percentage of 67 . 13 \%.\\ \hline

    \end{tabular}
\end{table*}

\begin{table*}[ht]
    \centering
    \renewcommand{\arraystretch}{1.2} 
    
    \begin{tabular}{l p{12cm}} 
        \hline \hline
        Origin & it should be described that the \textcolor[HTML]{db5079}{\textbf{ers}} signals derived from \textcolor[HTML]{db5079}{\textbf{perovskite}} layers cannot be observed because of their \textcolor[HTML]{db5079}{\textbf{pauli paramagnetism}} nature, resulting in low \textbf{ers} intensity, and because of the heavy-atom effects of \textcolor[HTML]{db5079}{\textbf{pb}} or \textbf{sn}, leading to short spin-lattice relaxation time and broad \textbf{ers} linewidths. \cite{chen2025operando}\\ \hline
        WordPiece & it should be described that the \textcolor[HTML]{db5079}{\textbf{esr}} signals derived from \textcolor[HTML]{db5079}{\textbf{perovsk-ite}} layers cannot be observed because of their \textcolor[HTML]{db5079}{\textbf{paul-i param-agne-tism}} nature , resulting in low \textbf{ers} intensity , and because of the heavy - atom effects of \textcolor[HTML]{db5079}{\textbf{p-b }} or \textbf{sn} , leading to short spin - lattice relaxation time and broad \textbf{ers} line-width-s. \\
        SAGE & it should be described that the \textcolor[HTML]{db5079}{\textbf{e-sr}} signals derived from \textcolor[HTML]{db5079}{\textbf{per-o-v-skite}} layers cannot be ob-served because of their \textcolor[HTML]{db5079}{\textbf{paul-i param-agnetism}} nature , resulting in low \textbf{e-sr} intens-ity , and because of the heavy - atom effects of \textcolor[HTML]{db5079}{\textbf{p-b }} or \textbf{sn} , leading to short spin - lattice relaxation time and broad \textbf{e-sr} linewidth-s. \\
        PickyBPE & it should be described that the \textcolor[HTML]{db5079}{\textbf{es-r}} signals derived from \textcolor[HTML]{db5079}{\textbf{perovskite}} layers cannot be observed because of their \textcolor[HTML]{db5079}{\textbf{pa-uli param-agnetism}} nature , resulting in low \textbf{es-r} intensity , and because of the heavy - atom effects of \textcolor[HTML]{db5079}{\textbf{p-b }} or \textbf{sn} , leading to short spin - lattice relaxation time and broad \textbf{es-r} linewidths. \\
        MATTER (ours) & it should be described that the \textcolor[HTML]{db5079}{\textbf{esr}} signals derived from \textcolor[HTML]{db5079}{\textbf{perovskite}} layers cannot be observed because of their \textcolor[HTML]{db5079}{\textbf{pauli paramagnetism}} nature , resulting in low \textbf{esr} intensity , and because of the heavy - atom effects of \textcolor[HTML]{db5079}{\textbf{pb}} or \textbf{sn} , leading to short spin - lattice relaxation time and broad \textbf{esr} linewidths. \\ \hline
         \hline
        Origin & in this study, the \textbf{nanoparticles} of the zinc and cobalt \textcolor[HTML]{db5079}{\textbf{imidazolate}} framework ( \textcolor[HTML]{db5079}{\textbf{znco}}-\textcolor[HTML]{db5079}{\textbf{zif}} ) were synthesized and directly incorporated into \textcolor[HTML]{db5079}{\textbf{polylactide}} ( \textcolor[HTML]{db5079}{\textbf{pla}} ) to prepare \textbf{pla / znco-zif fibrous} membranes through \textbf{electrospinning} methodology. \cite{deng2024incorporating} \\ \hline
        WordPiece &  in this study , the \textbf{nanoparticles} of the zinc and cobalt \textcolor[HTML]{db5079}{\textbf{im-ida-zol-ate}} framework ( \textcolor[HTML]{db5079}{\textbf{zn-co}} - \textcolor[HTML]{db5079}{\textbf{zi-f}} ) were synthesized and directly incorporated into \textcolor[HTML]{db5079}{\textbf{poly-lact-ide}} ( \textcolor[HTML]{db5079}{\textbf{pl-a}} ) to prepare \textbf{pla / zn-co - zi-f fibrous } membranes through \textbf{electros-pin-ning} methodology. \\
        SAGE & in this study , the \textbf{nanoparticles} of the zinc and cobalt \textcolor[HTML]{db5079}{\textbf{imid-azol-ate}} framework ( \textcolor[HTML]{db5079}{\textbf{z-nc-o}} - \textcolor[HTML]{db5079}{\textbf{z-if}} ) were synthesized and directly incorporated into \textcolor[HTML]{db5079}{\textbf{polyl-act-ide}} ( \textcolor[HTML]{db5079}{\textbf{pl-a}} ) to prepare \textbf{pl-a / z-nc-o - z-if fibrous} membranes through \textbf{electrospinning} methodology. \\
        PickyBPE & in this study, the \textbf{nanoparticles} of the zinc and cobalt \textcolor[HTML]{db5079}{\textbf{imid-az-olate}} framework ( \textcolor[HTML]{db5079}{\textbf{z-n-co}} - \textcolor[HTML]{db5079}{\textbf{z-if}} ) were synthesized and directly incorporated into \textcolor[HTML]{db5079}{\textbf{pol-yl-actide}} ( \textcolor[HTML]{db5079}{\textbf{pl-a}} ) to prepare \textbf{pl-a / z-n-co - z-if fibrous} membranes through \textbf{electrospinning} methodology. \\
        MATTER (ours) & in this study , the \textbf{nanoparticles} of the zinc and cobalt \textcolor[HTML]{db5079}{\textbf{imidazol-ate}} framework ( \textcolor[HTML]{db5079}{\textbf{zn-co}} - \textcolor[HTML]{db5079}{\textbf{zif}} ) were synthesized and directly incorporated into \textcolor[HTML]{db5079}{\textbf{polylactide}} ( \textcolor[HTML]{db5079}{\textbf{pla}} ) to prepare \textbf{pla / zn-co - zif fibrous} membranes through \textbf{electrospinning} methodology. \\ \hline

    \end{tabular}
    \caption{\textbf{Boldface} and \textcolor[HTML]{db5079}{\textbf{pink}} concepts are important material concepts extracted using MatDetector. \textbf{Boldface} concepts are correctly tokenized in both the baseline and our method, indicating no issues. In contrast, \textcolor[HTML]{db5079}{\textbf{pink}} concepts are highly important but are often split into unrelated subwords or overtokenized in conventional tokenization. However, as shown in this table, our method, MATTER, effectively prevents the overtokenization of important material concepts, preserving their semantic integrity.}
    \label{tab:tokenization_results_real}
\end{table*}

\begin{table*}[ht]
\centering
\arraybackslash
\fontsize{10}{10.2}\selectfont 
\setlength{\tabcolsep}{3pt} 
\renewcommand{\arraystretch}{1.3} 
\resizebox{\textwidth}{!}{%
\begin{tabular}{c|cll|cll|cll}
\hline
\hline
\textbf{\begin{tabular}[c]{@{}c@{}}Tokenization\\ Method\end{tabular}} & \textbf{Concept}                                                                                                                   & \multicolumn{1}{c}{\textbf{\begin{tabular}[c]{@{}c@{}}Word Embedding \\ 5-NN\end{tabular}}} & \multicolumn{1}{c|}{\textbf{Sim.}} & \textbf{Formula}                                                            & \multicolumn{1}{c}{\textbf{\begin{tabular}[c]{@{}c@{}}Word Embedding \\ 5-NN\end{tabular}}} & \multicolumn{1}{c|}{\textbf{Sim.}} & \textbf{Abbr}                                                            & \multicolumn{1}{c}{\textbf{\begin{tabular}[c]{@{}c@{}}Word Embedding \\ 5-NN\end{tabular}}} & \multicolumn{1}{c}{\textbf{Sim.}} \\ \hline
\multirow{5}{*}{\textbf{WordPiece}}                                    & \multirow{5}{*}{\begin{tabular}[c]{@{}c@{}}germanium\\ (german-ium)\end{tabular}}                                                  & agilent                                                                                     & 90.6                               & \multirow{5}{*}{\begin{tabular}[c]{@{}c@{}}PbI2\\ (pib-2)\end{tabular}}     & nowak                                                                                       & 81.8                               & \multirow{5}{*}{\begin{tabular}[c]{@{}c@{}}LFP\\ (lf-p)\end{tabular}}    & inlet                                                                                       & 95.5                              \\
                                                                       &                                                                                                                                    & fri                                                                                         & 85.9                               &                                                                             & 10c                                                                                         & 81.8                               &                                                                          & chattopadhyay                                                                               & 93.7                              \\
                                                                       &                                                                                                                                    & stephan                                                                                     & 85.3                               &                                                                             & -agnetically                                                                                & 81.0                               &                                                                          & 1263.0                                                                                      & 92.7                              \\
                                                                       &                                                                                                                                    & valley                                                                                      & 85.2                               &                                                                             & colouring                                                                                   & 79.6                               &                                                                          & foreland                                                                                    & 92.7                              \\
                                                                       &                                                                                                                                    & galvanic                                                                                    & 86.1                               &                                                                             & \textbf{quasicrystal}                                                                       & 79.6                               &                                                                          & -rink                                                                                       & 92.6                              \\ \hline
\multirow{5}{*}{\textbf{SAGE}}                                         & \multirow{5}{*}{\begin{tabular}[c]{@{}c@{}}germanium\\ (german-ium)\end{tabular}}                                                  & lot                                                                                         & 83.0                               & \multirow{5}{*}{\begin{tabular}[c]{@{}c@{}}PbI2\\ (p-bi-2)\end{tabular}}    & -gen                                                                                        & 43.9                               & \multirow{5}{*}{\begin{tabular}[c]{@{}c@{}}LFP\\ (lf-p)\end{tabular}}    & occupation                                                                                  & 95.8                              \\
                                                                       &                                                                                                                                    & segregation                                                                                 & 82.8                               &                                                                             & pounds                                                                                      & 43.6                               &                                                                          & multiphonon                                                                                 & 95.2                              \\
                                                                       &                                                                                                                                    & 100                                                                                         & 82.9                               &                                                                             & -pt                                                                                         & 43.5                               &                                                                          & -l                                                                                          & 95.2                              \\
                                                                       &                                                                                                                                    & segregation                                                                                 & 82.9                               &                                                                             & -uck                                                                                        & 43.2                               &                                                                          & -circ                                                                                       & 95.1                              \\
                                                                       &                                                                                                                                    & agi                                                                                         & 82.0                               &                                                                             & -8                                                                                          & 43.2                               &                                                                          & multiphonon                                                                                 & 95.1                              \\ \hline
\multirow{5}{*}{\textbf{PickyBPE}}                                     & \multirow{5}{*}{\begin{tabular}[c]{@{}c@{}}germanium\\ (g-erman-ium)\end{tabular}}                                                 & lot                                                                                         & 83.0                               & \multirow{5}{*}{\begin{tabular}[c]{@{}c@{}}PbI2\\ (p-bi-2)\end{tabular}}    & gaussian                                                                                    & 63.1                               & \multirow{5}{*}{\begin{tabular}[c]{@{}c@{}}LFP\\ (l-f-p)\end{tabular}}   & her,                                                                                        & 75.8                              \\
                                                                       &                                                                                                                                    & segregation                                                                                 & 82.8                               &                                                                             & p                                                                                           & 62.8                               &                                                                          & consideration                                                                               & 75.0                              \\
                                                                       &                                                                                                                                    & \textbf{-inov}                                                                              & 81.3                               &                                                                             & total                                                                                       & 62.3                               &                                                                          & -ermany                                                                                     & 75.0                              \\
                                                                       &                                                                                                                                    & -w,                                                                                         & 81.1                               &                                                                             & -'s                                                                                         & 62.0                               &                                                                          & -sd102                                                                                      & 75.0                              \\
                                                                       &                                                                                                                                    & compatibility                                                                               & 81.0                               &                                                                             & -ories                                                                                      & 61.0                               &                                                                          & \{{[}\}40\{{]}\}                                                                            & 75.0                              \\ \hline
\multirow{5}{*}{\textbf{MATTER (ours)}}                                & \multirow{5}{*}{\begin{tabular}[c]{@{}c@{}}germanium\\ (germanium)\end{tabular}}                                                   & \textbf{dithiocarbamate}                                                                    & 81.5                               & \multirow{5}{*}{\begin{tabular}[c]{@{}c@{}}PbI2\\ (pbi2)\end{tabular}}      & \textbf{pb5}                                                                                & 89.9                               & \multirow{5}{*}{\begin{tabular}[c]{@{}c@{}}LFP\\ (lfp)\end{tabular}}     & \textbf{zrf7}                                                                               & 90.9                              \\
                                                                       &                                                                                                                                    & \textbf{monium}                                                                             & 81.4                               &                                                                             & \textbf{pbf2}                                                                               & 89.2                               &                                                                          & \textbf{acyclohex}                                                                          & 90.8                              \\
                                                                       &                                                                                                                                    & -orib                                                                                       & 81.3                               &                                                                             & \textbf{-anesulfonic}                                                                       & 88.9                               &                                                                          & \textbf{dodecane}                                                                           & 90.0                              \\
                                                                       &                                                                                                                                    & \textbf{borohydride}                                                                        & 81.2                               &                                                                             & \textbf{-ob2o3}                                                                             & 88.7                               &                                                                          & \textbf{-acyclohex}                                                                         & 90.0                              \\
                                                                       &                                                                                                                                    & \textbf{-stannyl}                                                                           & 81.2                               &                                                                             & \textbf{-dithiocarbamate}                                                                   & 88.5                               &                                                                          & \textbf{-azobenzene}                                                                        & 90.0                              \\ \hline
\textbf{\begin{tabular}[c]{@{}c@{}}Tokenization\\ Method\end{tabular}} & \textbf{Concept}                                                                                                                   & \multicolumn{1}{c}{\textbf{\begin{tabular}[c]{@{}c@{}}Word Embedding \\ 5-NN\end{tabular}}} & \multicolumn{1}{c|}{\textbf{Sim.}} & \textbf{Formula}                                                            & \multicolumn{1}{c}{\textbf{\begin{tabular}[c]{@{}c@{}}Word Embedding \\ 5-NN\end{tabular}}} & \multicolumn{1}{c|}{\textbf{Sim.}} & \textbf{Abbr}                                                            & \multicolumn{1}{c}{\textbf{\begin{tabular}[c]{@{}c@{}}Word Embedding \\ 5-NN\end{tabular}}} & \multicolumn{1}{c}{\textbf{Sim.}} \\ \hline
\multirow{5}{*}{\textbf{WordPiece}}                                    & \multirow{5}{*}{\begin{tabular}[c]{@{}c@{}}ethylenediaminetetra-acetic acid\\ (ethylenedi-amine-tetr-aa-ce-tic acid)\end{tabular}} & -oreg                                                                                       & 92.4                               & \multirow{5}{*}{\begin{tabular}[c]{@{}c@{}}BaSo4\\ (bas-o-4)\end{tabular}}  & \textbf{bas}                                                                                & 91.3                               & \multirow{5}{*}{\begin{tabular}[c]{@{}c@{}}PDMS\\ (pd-ms)\end{tabular}}  & -ms                                                                                         & 87.6                              \\
                                                                       &                                                                                                                                    & \textbf{sulphates}                                                                          & 92.3                               &                                                                             & \textbf{-o4}                                                                                & 87.3                               &                                                                          & pd                                                                                          & 82.9                              \\
                                                                       &                                                                                                                                    & consistence                                                                                 & 92.2                               &                                                                             & \textbf{adopts}                                                                             & 87.0                               &                                                                          & drilled                                                                                     & 75.2                              \\
                                                                       &                                                                                                                                    & crop                                                                                        & 92.1                               &                                                                             & somehow                                                                                     & 85.3                               &                                                                          & gilbert                                                                                     & 75.1                              \\
                                                                       &                                                                                                                                    & -ulos                                                                                       & 92.1                               &                                                                             & \textbf{reflect}                                                                            & 85.1                               &                                                                          & connect                                                                                     & 74.3                              \\ \hline
\multirow{5}{*}{\textbf{SAGE}}                                         & \multirow{5}{*}{\begin{tabular}[c]{@{}c@{}}ethylenediaminetetra-acetic acid\\ (ethylenedi-amine-tetr-aa-ce-tic acid)\end{tabular}} & \textbf{-ilent}                                                                             & 94.8                               & \multirow{5}{*}{\begin{tabular}[c]{@{}c@{}}BaSo4\\ (bas-o-4)\end{tabular}}  & bas                                                                                         & 85.2                               & \multirow{5}{*}{\begin{tabular}[c]{@{}c@{}}PDMS\\ (pd-ms)\end{tabular}}  & p                                                                                           & 93.1                              \\
                                                                       &                                                                                                                                    & \textbf{athermal}                                                                           & 94.7                               &                                                                             & \textbf{nanobelts}                                                                          & 75.3                               &                                                                          & others                                                                                      & 91.5                              \\
                                                                       &                                                                                                                                    & -stoichi                                                                                    & 94.6                               &                                                                             & interv                                                                                      & 75.2                               &                                                                          & heas                                                                                        & 91.4                              \\
                                                                       &                                                                                                                                    & \textbf{thermogravimetric}                                                                  & 94.6                               &                                                                             & ).(                                                                                         & 75.1                               &                                                                          & ellips                                                                                      & 91.3                              \\
                                                                       &                                                                                                                                    & -true                                                                                       & 94.5                               &                                                                             & -rino                                                                                       & 74.8                               &                                                                          & -rac                                                                                        & 91.2                              \\ \hline
\multirow{5}{*}{\textbf{PickyBPE}}                                     & \multirow{5}{*}{ethyl-ened-i-amin-et-etra-acetic acid}                                                                             & contrast,                                                                                   & 89.6                               & \multirow{5}{*}{\begin{tabular}[c]{@{}c@{}}BaSo4\\ (b-as-o-4)\end{tabular}} & sliding                                                                                     & 71.6                               & \multirow{5}{*}{\begin{tabular}[c]{@{}c@{}}PDMS\\ (p-d-ms)\end{tabular}} & \textbf{ppe}                                                                                & 76.5                              \\
                                                                       &                                                                                                                                    & represents                                                                                  & 89.2                               &                                                                             & charged                                                                                     & 69.4                               &                                                                          & );                                                                                          & 69.9                              \\
                                                                       &                                                                                                                                    & \textbf{sophistic}                                                                          & 89.1                               &                                                                             & -adi                                                                                        & 67.7                               &                                                                          & \textbf{prem}                                                                               & 69.5                              \\
                                                                       &                                                                                                                                    & \textbf{zn(II)}                                                                             & 89.1                               &                                                                             & 2013.0                                                                                      & 60.2                               &                                                                          & inductively                                                                                 & 66.6                              \\
                                                                       &                                                                                                                                    & distribution                                                                                & 89.0                               &                                                                             & \textbf{mocvd}                                                                              & 59.8                               &                                                                          & bat                                                                                         & 66.6                              \\ \hline
\multirow{5}{*}{\textbf{MATTER (ours)}}                                & \multirow{5}{*}{ethylenediaminetetra-acetic acid}                                                                                  & \textbf{ethylenediaminetetra}                                                               & 93.7                               & \multirow{5}{*}{\begin{tabular}[c]{@{}c@{}}BaSo4\\ (baso40\end{tabular}}    & \textbf{bas}                                                                                & 91.4                               & \multirow{5}{*}{\begin{tabular}[c]{@{}c@{}}PDMS\\ (pdms)\end{tabular}}   & \textbf{perfluoroalkyl}                                                                     & 85.7                              \\
                                                                       &                                                                                                                                    & \textbf{-acetic}                                                                            & 93.6                               &                                                                             & \textbf{-o4}                                                                                & 87.5                               &                                                                          & \textbf{trimethoxysilyl}                                                                    & 85.6                              \\
                                                                       &                                                                                                                                    & \textbf{-oxycarb}                                                                           & 91.8                               &                                                                             & \textbf{bast}                                                                               & 84.4                               &                                                                          & \textbf{-yloxy}                                                                             & 85.5                              \\
                                                                       &                                                                                                                                    & \textbf{-sulfanyl}                                                                          & 91.7                               &                                                                             & \textbf{cyclohexyl}                                                                         & 82.0                               &                                                                          & \textbf{-obenzoic}                                                                          & 85.4                              \\
                                                                       &                                                                                                                                    & \textbf{agre}                                                                               & 91.6                               &                                                                             & \textbf{-cyclopentadienyl}                                                                  & 81.9                               &                                                                          & \textbf{borohyd}                                                                            & 85.4                              \\ \hline
 \hline
\end{tabular}
}
  \caption{Comparison of subword embedding averaging results across different tokenization methods, including WordPiece, SAGE, PickyBPE, and our proposed method, MATTER. The table presents the five nearest neighbor words based on subword embedding averages for each method, illustrating how different tokenization strategies impact semantic similarity in word embeddings. The similarity scores (Sim.) indicate the relevance of the nearest neighbors to the target material concept. \textbf{Boldface} highlights words that are directly related to materials.}
  \label{tab:sim top5}
\end{table*}

\begin{table*}[ht]
\centering
\arraybackslash

\resizebox{\textwidth}{!}{%
\begin{tabular}{cccccccccc}
\toprule
\multicolumn{2}{c}{} & \multicolumn{8}{c}{MatSci-NLP} \\ \cmidrule{3-10} 
\multicolumn{2}{c}{\multirow{-2}{*}{MatDetector (ours)}} & NER & RC & EAE & PC & SAR & SC & SF & Overall \\ \midrule 
 & Micro-F1 & 76.6 & 80.9 & 48.5 & 73.1 & 81.9 & 90.0 & 57.4 & 72.6 \\
\multirow{-2}{*}{w/o  material knowledge} & Macro-F1 & 56.1 & 58.5 & 29.4 & 58.9 & 74.6 & 60.3 & 32.6 & 52.9 \\ \midrule  \midrule 
 & Micro-F1 & 76.4 & 78.6 & 47.4 & 74.1 & 79.5 & 91.0 & 61.2 & 72.6 \\
\multirow{-2}{*}{0.1} & Macro-F1 & 54.3 & 54.4 & 32.6 & 69.5 & 62.7 & 61.1 & 37.0 & {   53.1} \\ \midrule 
 & Micro-F1 & 78.3 & 78.7 & 49.7 & { 74.5} & 76.2 & \textbf{91.2} & 60.4 & 72.7 \\
\multirow{-2}{*}{0.2} & Macro-F1 & { 58.5} & 53.1 & 30.2 & 68.8 & 63.3 & 58.0 & 36.7 & {   52.6} \\ \midrule 
 & Micro-F1 & 78.5 & 80.2 & 51.2 & \underline{76.6} & 73.9 & 91.5 & \textbf{62.7} & 73.5 \\
\multirow{-2}{*}{0.3} & Macro-F1 & 56.2 & 55.7 & 35.7 & 69.2 & 58.9 & 62.5 & \textbf{41.3} & 54.2 \\ \midrule 
 & Micro-F1 & 75.4 & 80.7 & 52.9 & 73.3 & 77.2 & 90.6 & 59.1 & 72.7 \\
\multirow{-2}{*}{0.4} & Macro-F1 & 54.5 & 56.5 & 33.4 & 68.4 & 59.9 & 63.5 & 35.0 & 53.0 \\ \midrule 
 & Micro-F1 & {   78.7} & {   82.4} & 53.0 & 74.2 & 76.2 & 90.8 & 60.2 & 73.6 \\
\multirow{-2}{*}{0.5} & Macro-F1 & 55.2 & 58.4 & 32.6 & {   66.5} & 63.9 & 61.5 & 33.5 & 53.1 \\ \midrule 
 & Micro-F1 & 77.8 & \underline{83.6} & {   49.6} & \textbf{77.3} & {   78.3} & \textbf{91.2} & 61.2 & 74.1 \\
\multirow{-2}{*}{0.6} & Macro-F1 & 57.2 & \textbf{60.7} & {   30.7} & 68.9 & 63.1 & 59.3 & 38.0 & 54.0 \\ \midrule 
 & Micro-F1 & 78.3 & 81.5 & 48.3 & 74.5 & \underline{82.4} & 90.9 & 59.8 & {   73.7} \\
\multirow{-2}{*}{0.7} & Macro-F1 & 56.7 & 58.0 & 34.9 & 69.7 & {   60.0} & 59.6 & 37.4 & 53.8 \\ \midrule 
 & Micro-F1 & {   76.9} & 80.4 & \textbf{54.0} & 75.2 & 80.4 & {   91.1} & 61.5 & \underline{74.2} \\
\multirow{-2}{*}{0.8} & Macro-F1 & {   54.9} & {   55.7} & 37.0 & {   \textbf{71.4}} & \underline{74.7} & 59.5 & 37.8 & {   55.9} \\ \midrule 
 & Micro-F1 & \underline{79.0} & {81.3} & {   53.1} & {   74.2} & {   {78.6}} & {90.2} & {   60.8} & {73.9} \\
\multirow{-2}{*}{0.9} & Macro-F1 & {\underline{58.8}} & 59.0 & {   \textbf{37.3}} & \underline{69.5} & {64.1} & {   \textbf{66.4}} & 37.7 & {\underline{56.1}} \\ \midrule 
\rowcolor[HTML]{E9EBF5} 
\cellcolor[HTML]{E9EBF5} & Micro-F1 & \textbf{80.0} & \textbf{83.8} & \underline{53.1} & {73.7} & \textbf{85.5} & \textbf{91.2} & {   \underline{61.9}} & \textbf{75.6} \\
\rowcolor[HTML]{E9EBF5} 
\multirow{-2}{*}{\cellcolor[HTML]{E9EBF5}1.0} & Macro-F1 & \textbf{59.3} & \underline{59.1} & \underline{36.9} & 67.6 & \textbf{79.3} & \underline{64.9} & {   38.0} & \textbf{57.9} \\ \midrule 
 & Micro-F1 & {   77.3} & 79.2 & 52.1 & 75.1 & 75.7 & 91.1 & 61.1 & 73.1 \\
\multirow{-2}{*}{2.0} & Macro-F1 & 55.7 & 55.9 & 36.6 & 66.6 & {   62.6} & 60.1 & {   38.0} & 53.7 \\ \midrule 
 & Micro-F1 & 76.1 & 79.2 & 50.5 & 71.6 & 77.9 & 90.2 & {   61.5} & 72.4 \\
\multirow{-2}{*}{3.0} & Macro-F1 & 54.2 & 57.6 & 34.2 & 65.9 & 63.8 & 59.7 & \underline{38.6} & 53.4 \\ \bottomrule

\end{tabular}
}

  \caption{Specific numerical results of MatDetector's Macro-F1 and Micro-F1 scores across different $\lambda$ values. }
  \label{tab:ab1}
\end{table*}

\begin{table*}[ht]
\centering
\arraybackslash
\resizebox{\textwidth}{!}{%
\begin{tabular}{cccccccccc}
\toprule
\multicolumn{2}{c}{{\color[HTML]{000000} }} & \multicolumn{8}{c}{\cellcolor[HTML]{FFFFFF}{\color[HTML]{000000} MatSci-NLP}} \\ \cmidrule{3-10} 
\multicolumn{2}{c}{\multirow{-2}{*}{{\color[HTML]{000000} \begin{tabular}[c]{@{}c@{}}ChemDataExtractor\\ \cite{swain2016chemdataextractor}\end{tabular}}}} & \cellcolor[HTML]{FFFFFF}{\color[HTML]{000000} NER} & \cellcolor[HTML]{FFFFFF}{\color[HTML]{000000} RC} & \cellcolor[HTML]{FFFFFF}{\color[HTML]{000000} EAE} & \cellcolor[HTML]{FFFFFF}{\color[HTML]{000000} PC} & \cellcolor[HTML]{FFFFFF}{\color[HTML]{000000} SAR} & \cellcolor[HTML]{FFFFFF}{\color[HTML]{000000} SC} & \cellcolor[HTML]{FFFFFF}{\color[HTML]{000000} SF} & \cellcolor[HTML]{FFFFFF}{\color[HTML]{000000} Overall} \\ \midrule 
{\color[HTML]{000000} } & \cellcolor[HTML]{FFFFFF}{\color[HTML]{000000} Micro-F1} & {\color[HTML]{000000} 76.6} & {\color[HTML]{000000} 80.9} & {\color[HTML]{000000} 48.5} & {\color[HTML]{000000} 73.1} & {\color[HTML]{000000} 81.9} & {\color[HTML]{000000} 90.0} & {\color[HTML]{000000} 57.4} & {\color[HTML]{000000} 72.6} \\
\multirow{-2}{*}{{\color[HTML]{000000} w/o  material knowledge}} & \cellcolor[HTML]{FFFFFF}{\color[HTML]{000000} Macro-F1} & {\color[HTML]{000000} 56.1} & {\color[HTML]{000000} 58.5} & {\color[HTML]{000000} 29.4} & {\color[HTML]{000000} 58.9} & {\color[HTML]{000000} \textbf{74.6}} & {\color[HTML]{000000} 60.3} & {\color[HTML]{000000} 32.6} & {\color[HTML]{000000} 52.9} \\ \midrule \midrule 
{\color[HTML]{000000} } & \cellcolor[HTML]{FFFFFF}{\color[HTML]{000000} Micro-F1} & {\color[HTML]{000000} 75.5} & {\color[HTML]{000000} 81.0} & {\color[HTML]{000000} 52.7} & {\color[HTML]{000000} 72.8} & {\color[HTML]{000000} 77.9} & {\color[HTML]{000000} 90.4} & {\color[HTML]{000000} 55.8} & {\color[HTML]{000000} 72.3} \\
\multirow{-2}{*}{{\color[HTML]{000000} 0.1}} & \cellcolor[HTML]{FFFFFF}{\color[HTML]{000000} Macro-F1} & {\color[HTML]{000000} 52.4} & {\color[HTML]{000000} 61.1} & {\color[HTML]{000000} 34.4} & {\color[HTML]{000000} 63.0} & {\color[HTML]{000000} 66.6} & {\color[HTML]{000000} 62.2} & {\color[HTML]{000000} 31.9} & {\color[HTML]{000000} 53.1} \\ \midrule 
{\color[HTML]{000000} } & \cellcolor[HTML]{FFFFFF}{\color[HTML]{000000} Micro-F1} & {\color[HTML]{000000} 76.4} & {\color[HTML]{000000} 83.2} & {\color[HTML]{000000} 52.1} & {\color[HTML]{000000} 70.5} & {\color[HTML]{000000} 76.7} & {\color[HTML]{000000} 91.0} & {\color[HTML]{000000} 58.6} & {\color[HTML]{000000} 72.6} \\
\multirow{-2}{*}{{\color[HTML]{000000} 0.2}} & \cellcolor[HTML]{FFFFFF}{\color[HTML]{000000} Macro-F1} & {\color[HTML]{000000} \underline{56.3}} & {\color[HTML]{000000} 61.0} & {\color[HTML]{000000} 32.8} & {\color[HTML]{000000} 64.8} & {\color[HTML]{000000} 63.8} & {\color[HTML]{000000} 60.3} & {\color[HTML]{000000} 34.2} & {\color[HTML]{000000} 53.3} \\ \midrule 
{\color[HTML]{000000} } & \cellcolor[HTML]{FFFFFF}{\color[HTML]{000000} Micro-F1} & {\color[HTML]{000000} 75.4} & {\color[HTML]{000000} 82.3} & {\color[HTML]{000000} 52.3} & {\color[HTML]{000000} 73.1} & {\color[HTML]{000000} 76.4} & {\color[HTML]{000000} 90.4} & {\color[HTML]{000000} 56.8} & {\color[HTML]{000000} 72.4} \\
\multirow{-2}{*}{{\color[HTML]{000000} 0.3}} & \cellcolor[HTML]{FFFFFF}{\color[HTML]{000000} Macro-F1} & {\color[HTML]{000000} 53.2} & {\color[HTML]{000000} 61.0} & {\color[HTML]{000000} 29.8} & {\color[HTML]{000000} 65.2} & {\color[HTML]{000000} 64.4} & {\color[HTML]{000000} 58.9} & {\color[HTML]{000000} 33.0} & {\color[HTML]{000000} 52.2} \\ \midrule 
{\color[HTML]{000000} } & \cellcolor[HTML]{FFFFFF}{\color[HTML]{000000} Micro-F1} & {\color[HTML]{000000} 73.4} & {\color[HTML]{000000} 84.0} & {\color[HTML]{000000} \textbf{55.1}} & {\color[HTML]{000000} 71.9} & {\color[HTML]{000000} 79.5} & {\color[HTML]{000000} 90.6} & {\color[HTML]{000000} 57.1} & {\color[HTML]{000000} 73.1} \\
\multirow{-2}{*}{{\color[HTML]{000000} 0.4}} & \cellcolor[HTML]{FFFFFF}{\color[HTML]{000000} Macro-F1} & {\color[HTML]{000000} 51.4} & {\color[HTML]{000000} 58.4} & {\color[HTML]{000000} \textbf{37.9}} & {\color[HTML]{000000} \textbf{68.8}} & {\color[HTML]{000000} 66.6} & {\color[HTML]{000000} 60.0} & {\color[HTML]{000000} 30.5} & {\color[HTML]{000000} 53.4} \\ \midrule 
{\color[HTML]{000000} } & \cellcolor[HTML]{FFFFFF}{\color[HTML]{000000} Micro-F1} & {\color[HTML]{000000} \underline{77.0}} & {\color[HTML]{000000} 82.3} & {\color[HTML]{000000} 53.8} & {\color[HTML]{000000} 72.2} & {\color[HTML]{000000} 79.8} & {\color[HTML]{000000} 91.0} & {\color[HTML]{000000} 57.7} & {\color[HTML]{000000} 73.4} \\
\multirow{-2}{*}{{\color[HTML]{000000} 0.5}} & \cellcolor[HTML]{FFFFFF}{\color[HTML]{000000} Macro-F1} & {\color[HTML]{000000} \textbf{56.4}} & {\color[HTML]{000000} 61.3} & {\color[HTML]{000000} 35.8} & {\color[HTML]{000000} \underline{67.7}} & {\color[HTML]{000000} 62.0} & {\color[HTML]{000000} 61.8} & {\color[HTML]{000000} 30.8} & {\color[HTML]{000000} 53.7} \\ \midrule 
{\color[HTML]{000000} } & \cellcolor[HTML]{FFFFFF}{\color[HTML]{000000} Micro-F1} & {\color[HTML]{000000} 76.5} & {\color[HTML]{000000} \underline{84.1}} & {\color[HTML]{000000} \underline{54.1}} & {\color[HTML]{000000} 67.3} & {\color[HTML]{000000} 78.2} & {\color[HTML]{000000} \textbf{91.2}} & {\color[HTML]{000000} 57.8} & {\color[HTML]{000000} 72.7} \\
\multirow{-2}{*}{{\color[HTML]{000000} 0.6}} & \cellcolor[HTML]{FFFFFF}{\color[HTML]{000000} Macro-F1} & {\color[HTML]{000000} 55.1} & {\color[HTML]{000000} 61.8} & {\color[HTML]{000000} {   36.6}} & {\color[HTML]{000000} 59.6} & {\color[HTML]{000000} 66.2} & {\color[HTML]{000000} 61.9} & {\color[HTML]{000000} 33.6} & {\color[HTML]{000000} 53.5} \\ \midrule 
{\color[HTML]{000000} } & \cellcolor[HTML]{FFFFFF}{\color[HTML]{000000} Micro-F1} & {\color[HTML]{000000} 75.4} & {\color[HTML]{000000} 82.4} & {\color[HTML]{000000} 52.5} & {\color[HTML]{000000} 71.8} & {\color[HTML]{000000} \textbf{82.3}} & {\color[HTML]{000000} 90.2} & {\color[HTML]{000000} 58.0} & {\color[HTML]{000000} 73.2} \\
\multirow{-2}{*}{{\color[HTML]{000000} 0.7}} & \cellcolor[HTML]{FFFFFF}{\color[HTML]{000000} Macro-F1} & {\color[HTML]{000000} 54.5} & {\color[HTML]{000000} 60.7} & {\color[HTML]{000000} 33.3} & {\color[HTML]{000000} 65.8} & {\color[HTML]{000000} 68.1} & {\color[HTML]{000000} 63.2} & {\color[HTML]{000000} \textbf{37.9}} & {\color[HTML]{000000} 54.8} \\ \midrule 
{\color[HTML]{000000} } & \cellcolor[HTML]{FFFFFF}{\color[HTML]{000000} Micro-F1} & {\color[HTML]{000000} 76.0} & {\color[HTML]{000000} 84.0} & {\color[HTML]{000000} 53.4} & {\color[HTML]{000000} 71.1} & {\color[HTML]{000000} 78.1} & {\color[HTML]{000000} {   91.1}} & {\color[HTML]{000000} 58.1} & {\color[HTML]{000000} 73.1} \\
\multirow{-2}{*}{{\color[HTML]{000000} 0.8}} & \cellcolor[HTML]{FFFFFF}{\color[HTML]{000000} Macro-F1} & {\color[HTML]{000000} 55.5} & {\color[HTML]{000000} \underline{62.8}} & {\color[HTML]{000000} 34.7} & {\color[HTML]{000000} 64.7} & {\color[HTML]{000000} 65.8} & {\color[HTML]{000000} \textbf{65.0}} & {\color[HTML]{000000} 35.5} & {\color[HTML]{000000} \underline{54.9}} \\ \hline
{\color[HTML]{000000} } & \cellcolor[HTML]{FFFFFF}{\color[HTML]{000000} Micro-F1} & {\color[HTML]{000000} 75.6} & {\color[HTML]{000000} 81.9} & {\color[HTML]{000000} 52.8} & {\color[HTML]{000000} \underline{73.3}} & {\color[HTML]{000000} \underline{82.0}} & {\color[HTML]{000000} 91.1} & {\color[HTML]{000000} 58.4} & \underline{\color[HTML]{000000} 73.6} \\
\multirow{-2}{*}{{\color[HTML]{000000} 0.9}} & \cellcolor[HTML]{FFFFFF}{\color[HTML]{000000} Macro-F1} & {\color[HTML]{000000} 54.1} & {\color[HTML]{000000} 59.0} & {\color[HTML]{000000} \underline{37.5}} & {\color[HTML]{000000} 67.5} & {\color[HTML]{000000} 65.5} & {\color[HTML]{000000} 58.0} & {\color[HTML]{000000} 35.7} & {\color[HTML]{000000} 53.9} \\ \midrule 
\rowcolor[HTML]{E9EBF5} 
\cellcolor[HTML]{E9EBF5}{\color[HTML]{000000} } & {\color[HTML]{000000} Micro-F1} & {\color[HTML]{000000} \textbf{77.1}} & {\color[HTML]{000000} 81.5} & {\color[HTML]{000000} 53.1} & {\color[HTML]{000000} \textbf{73.6}} & {\color[HTML]{000000} 80.6} & {\color[HTML]{000000} \textbf{91.2}} & {\color[HTML]{000000} \textbf{58.8}} & {\color[HTML]{000000} \textbf{73.7}} \\
\rowcolor[HTML]{E9EBF5} 
\multirow{-2}{*}{\cellcolor[HTML]{E9EBF5}{\color[HTML]{000000} 1.0}} & {\color[HTML]{000000} Macro-F1} & {\color[HTML]{000000} \textbf{56.4}} & {\color[HTML]{000000} 58.9} & {\color[HTML]{000000} 35.0} & {\color[HTML]{000000} 67.6} & {\color[HTML]{000000} 68.0} & {\color[HTML]{000000} \underline{64.8}} & {\color[HTML]{000000} \underline{35.6}} & {\color[HTML]{000000} \textbf{55.2}} \\  \midrule 
{\color[HTML]{000000} } & \cellcolor[HTML]{FFFFFF}{\color[HTML]{000000} Micro-F1} & {\color[HTML]{000000} \underline{77.0}} & {\color[HTML]{000000} \textbf{84.7}} & {\color[HTML]{000000} 52.3} & {\color[HTML]{000000} 69.4} & {\color[HTML]{000000} 80.7} & \underline{\color[HTML]{000000} 91.1} & {\color[HTML]{000000} 57.3} & {\color[HTML]{000000} 73.2} \\
\multirow{-2}{*}{{\color[HTML]{000000} 2.0}} & \cellcolor[HTML]{FFFFFF}{\color[HTML]{000000} Macro-F1} & {\color[HTML]{000000} 55.3} & {\color[HTML]{000000} \textbf{64.2}} & {\color[HTML]{000000} 34.1} & {\color[HTML]{000000} 64.3} & {\color[HTML]{000000} \underline{68.2}} & {\color[HTML]{000000} 60.3} & {\color[HTML]{000000} 35.4} & {\color[HTML]{000000} 54.5} \\ \midrule 
{\color[HTML]{000000} } & \cellcolor[HTML]{FFFFFF}{\color[HTML]{000000} Micro-F1} & {\color[HTML]{000000} 76.1} & {\color[HTML]{000000} 83.2} & {\color[HTML]{000000} 52.0} & {\color[HTML]{000000} 67.7} & {\color[HTML]{000000} 75.6} & {\color[HTML]{000000} 90.2} & {\color[HTML]{000000} \underline{58.7}} & {\color[HTML]{000000} 71.9} \\
\multirow{-2}{*}{{\color[HTML]{000000} 3.0}} & \cellcolor[HTML]{FFFFFF}{\color[HTML]{000000} Macro-F1} & {\color[HTML]{000000} 55.5} & {\color[HTML]{000000} 60.6} & {\color[HTML]{000000} 34.0} & {\color[HTML]{000000} 65.8} & {\color[HTML]{000000} 67.8} & {\color[HTML]{000000} 60.6} & {\color[HTML]{000000} 34.5} & {\color[HTML]{000000} 54.1}\\ \bottomrule 

\end{tabular}
}

  \caption{Specific numerical results of ChemDataExtractor's Macro-F1 and Micro-F1 scores across different $\lambda$ values. }
  \label{tab:ab2}
\end{table*}

\end{document}